%% file: main.tex
\documentclass{article}

\PassOptionsToPackage{numbers, compress}{natbib}

\usepackage[eandd, final]{neurips_2026}

\usepackage[utf8]{inputenc} % allow utf-8 input
\usepackage[T1]{fontenc}    % use 8-bit T1 fonts
\usepackage{hyperref}       % hyperlinks
\usepackage{url}            % simple URL typesetting
\usepackage{booktabs}       % professional-quality tables
\usepackage{amsfonts}       % blackboard math symbols
\usepackage{nicefrac}       % compact symbols for 1/2, etc.
\usepackage{microtype}      % microtypography
\usepackage{xcolor}         % colors
\usepackage{colortbl}

\newcommand*{\expertset}{URSA-expert-2026}

\usepackage{pdfpages}
\usepackage{multirow}
\usepackage[most]{tcolorbox}
\usepackage{xcolor}
\usepackage{caption}
\usepackage{csvsimple}
\usepackage{longtable}
\usepackage{seqsplit}
\usepackage{float}
\usepackage[section,above,below]{placeins}
\usepackage{rotating}
\usepackage{booktabs}
\definecolor{ursaGreen}{HTML}{82AC3D}
\definecolor{ursaAmber}{HTML}{D59733}
\definecolor{ursaCoral}{HTML}{D9686B}

\newtcolorbox{windowbox}[2][]{%
  enhanced,
  breakable,
  colback=white,
  colframe=black,
  boxrule=0.8pt,
  arc=2mm,
  left=3mm,right=3mm,top=2mm,bottom=2mm,
  fonttitle=\bfseries,
  coltitle=white,
  colbacktitle=black,
  title={#2},
  attach boxed title to top left={xshift=0mm,yshift=-1mm},
  boxed title style={sharp corners, boxrule=0pt, interior style={fill=black}},
  #1
}

\newtcolorbox{fieldbox}[2][]{%
  enhanced,
  breakable,
  colback=gray!3,
  colframe=black!25,
  boxrule=0.5pt,
  arc=1.5mm,
  left=2mm,right=2mm,top=1mm,bottom=1mm,
  fonttitle=\bfseries,
  title={#2},
  listing only,
  listing options={
    basicstyle=\ttfamily\footnotesize,
    columns=fullflexible,
    breaklines=true,
    showstringspaces=false,
  },
  #1
}

\definecolor{gold}{RGB}{255, 215, 0}
\definecolor{silver}{RGB}{192, 192, 192}
\definecolor{bronze}{RGB}{205, 127, 50}
\definecolor{emerald}{RGB}{21, 184, 166}
\definecolor{amber}{RGB}{251, 192, 36}
\definecolor{ruby}{RGB}{252, 113, 133}

\title{
URSA: Chemistry-Aware Benchmark for \\
Utilitarian Retrosynthesis Assessment
}

\author{%
  \bfseries Bogdan Zagribelnyy$^{1}$~\thanks{\texttt{bogdan@insilicomedicine.com}} \quad
  Ivan Ilin$^{1}$ \quad
  Nikita Bondarev$^{1}$ \quad
  Anton Morgunov$^{2}$ \\
  \bfseries Arkadii Lin$^{1}$ \quad
  Maksim Kuznetsov$^{3}$ \quad
  Rim Shayakhmetov$^{1}$ \\
  \bfseries Vladimir Aladinskiy$^{1}$ \quad
  Alex Aliper$^{1}$ \quad
  Alex Zhavoronkov$^{1,3,4}$ \\[0.6em]
  \small $^{1}$Insilico Medicine AI Limited, Masdar City, Abu Dhabi, UAE \\
  \small $^{2}$Independent researcher \\
  \small $^{3}$Insilico Medicine Canada Inc., Montreal, Quebec, Canada \\
  \small $^{4}$Insilico Medicine Hong Kong Ltd., Hong Kong SAR, China \\[0.3em]
}

\begin{document}

\maketitle

\begin{abstract}
Synthesis planning aiming to find pathways of reactions for a target molecule is one of the most important and challenging tasks in drug discovery. Recent progress has produced both specialized deep-learning retrosynthesis systems and general-purpose large language models, but objective comparison remains difficult due to the lack of flexible, chemically interpretable benchmarking protocols. In the current study, we are introducing the URSA (Utilitarian RetroSynthesis Assessment) evaluation framework that provides the opportunity to benchmark the synthetic routes not only from a formal perspective, such as convergence to commercially available starting materials, but also from a chemical plausibility perspective, mimicking the way expert chemists evaluate the reactions and routes. The study covers a comprehensive evaluation of both conventional end-to-end retrosynthesis solutions and LLMs for the synthesis planning task on a set of novel, diverse target molecules with undisclosed synthetic routes, which represent realistic tasks in the daily drug design routine. Although LLMs can support high-level strategic planning and even achieve SOTA performance in the OOD set, the Pareto frontier for estimated run cost and chemically plausible route solvability consists entirely of specialized retrosynthesis planners.
\end{abstract}

\input{text/main/intro}

\input{text/main/related_work}
\input{text/main/preliminaries}
\input{text/main/approach}
\input{text/main/experiments}

\input{text/main/results}

\input{text/main/limitations}
\input{text/main/conclusion}

\input{text/main/availability}

\input{text/main/funding}
\bibliography{bibliography}
\bibliographystyle{unsrtnat}

% APPENDIX

\newpage
\appendix
\onecolumn

\input{text/main/impact}
\input{text/appendix/Code_and_data_availablity}
\input{text/appendix/glossary}
\input{text/appendix/evaluated_systems}
\input{text/appendix/plausibility_bencmark_set}
\input{text/appendix/URSA-expert-2026}

\input{text/appendix/Sovability_benchmark}
\input{text/appendix/da_dca}
\input{text/appendix/MSTS_templates}
\input{text/appendix/uspto-190-ctitics}
\input{text/appendix/subtree_gen_example}
\input{text/appendix/ChemCensor}
\input{text/appendix/Synthegy}

\input{text/appendix/ranking_sensetivity}

\end{document}

%% file: text/main/intro.tex
\section{Introduction}
\label{sec:intro}

Computer-Aided/Assisted Synthesis Planning (CASP) aims to design synthetic pathways to target molecules, a long-standing grand challenge in organic chemistry~\cite{coley_grandchallenges_2026}. Since Corey and Wipke's early logic-centered systems~\cite{casp_1969}, most CASP methods have treated retrosynthesis as backward search over precursor trees: single-step models propose disconnections, and planners search for routes that terminate in available starting materials. Modern neural predictors and search algorithms~\cite{segler2018-solvability-mcts, askcos_2025, chen2020retrolearningretrosyntheticplanning, selfimproved_retro_2021, dreamretroer_2025, retrograph, grasp_2022, shee2025directmultistep} have greatly improved this navigability problem, but the dominant evaluation metric still asks a narrow question: "\textit{Did the planner find any path to stock?}"

This metric, often reported as solvability, has limited chemical meaning. Its value depends strongly on the definition of the starting-material stock, which varies from realistic commercial inventories to expansive virtual libraries. More importantly, stock termination does not imply that the individual transformations in the route are chemically plausible. Recent evaluations have shown that stock-terminated routes can contain chemically invalid transformations despite satisfying the formal search objective~\cite{tetko2024models-matter, morgunov2025procrustean-retrocast}. Thus, high stock-termination scores can reward graph connectivity while leaving unresolved the question a chemist actually cares about: "\textit{Is this route chemically sound?}"

Route-reproduction benchmarks such as PaRoutes provide a stronger signal by testing whether models recover known experimental syntheses~\cite{paroutes_2022}. However, exact reproduction is conservative: it cannot reward novel but plausible routes that differ from the reference synthesis. This leaves a measurement gap between two incomplete criteria. Stock termination can accept implausible novelty; route reproduction can reject plausible novelty. Recent work formalizes this distinction through the \texttt{Solv-N[task]} hierarchy, which separates syntactic (\texttt{N = 0}) and topological validity (\texttt{N = 1}) from higher-order requirements such as selectivity (\texttt{N = 2}) and experimental executability (\texttt{N = 3}) and from user-defined constraints \texttt{[task]} such as, e.g., termination to a specified stock of starting materials (SM), maximum allowed route length or restriction on the usage of specific reactions~\cite{Morgunov205Syntax}.

We introduce URSA (Utilitarian RetroSynthesis Assessment), a benchmark framework for evaluating retrosynthetic routes at the practically urgent Solv-1/Solv-2 boundary. URSA asks whether a proposed stock-terminated route advances beyond Solv-0/1 checks by satisfying Solv-2 chemical plausibility criteria at every reaction step. To support heterogeneous CASP systems, URSA uses RetroCast as a route-standardization layer~\cite{morgunov2025procrustean-retrocast}, then evaluates each standardized route with reaction-level plausibility checks.

The main contributions of this work are:

\begin{enumerate}
    \item We introduce URSA, a unified framework for evaluating the chemical plausibility of retrosynthetic routes generated for novel target molecules.
    \item We publish \texttt{Reaction plausibility benchmark dataset} collected by expert chemists to measure chemical plausibility discrimination power.
    \item We release \texttt{URSA-drugs\&clinicals-2026}, a diverse benchmark of approved drugs and clinical candidates designed to better reflect medicinal chemistry use cases, and expanded \texttt{URSA-drugs\&clinicals\&agrochemicals-2026}, a 333-target set that includes 33 agrochemicals and probes a broader chemical space.
    \item We benchmark modern synthesis planning tools, including conventional CASP systems and closed- and open-weight LLMs, under a route-level plausibility protocol.
\end{enumerate}

%% file: text/main/related_work.tex
\section{Related Work}
\label{sec:related_work}

\subsection{Benchmarking tools for synthesis planning}

Existing retrosynthesis benchmarks address different parts of the evaluation problem. PaRoutes introduced a large open dataset of experimentally reported routes and exposed the gap between stock termination and route reproduction~\cite{paroutes_2022}. Syntheseus broadened infrastructure support by making it easier to compose single-step retrosynthesis models with multistep planners such as MCTS and Retro*~\cite{chen2020retrolearningretrosyntheticplanning, maziarz2025syntheseus, segler2018-solvability-mcts}. RetroCast addressed the complementary standardization problem by converting heterogeneous CASP outputs into a common route representation, enabling reproducible scoring and qualitative inspection through SynthArena~\cite{morgunov2025procrustean-retrocast, SynthArena}. These tools improve comparability, but they do not by themselves provide an automated Solv-2 plausibility judge.

Recent work has begun to target this missing validity layer. RetroTrim proposes an ensemble of reaction scorers for hallucination filtering, but its current preprint does not release code or model weights for independent benchmarking~\cite{sadowski2025retrotrim-hallucinations}. LLM-based systems have also been proposed as synthesis evaluators or planners~\cite{xuan2025synthelite, Schwaller2026-synthegy, armstrong2025synthstrategyextractingformalizinglatent}, but their reliability on novel target molecules remains insufficiently established, and we identify several chemically problematic judgments in \autoref{app:synthefy_critics}. Reaction-procedure models such as MOSAIC~\cite{Li2026-MOSAIC} generate experimental protocols for supplied transformations, but procedure generation alone does not establish whether those transformations satisfy the Solv-2 selectivity constraints that precede route-level Solv-3 executability. ChemCensor instead evaluates reaction plausibility through reaction-center and functional-group precedent extracted from reported chemistry~\cite{zagribelnyy2026chemcensor}, making it a natural candidate for route-level Solv-2 benchmarking.

\subsection{Benchmarking target sets for synthesis planning}

Benchmarking sets also shape what kind of capability is measured. Widely used retrospective sets such as \texttt{USPTO-190}~\cite{chen2020retrolearningretrosyntheticplanning}, derived from \texttt{USPTO-full}~\cite{Lowe2017usptofull}, contain molecules with previously reported routes. They are useful for controlled retrospective evaluation, but they weakly reflect the prospective medicinal chemistry setting, where the target molecule may be novel and no reference synthesis is available. Recent analyses further show that such benchmarks can obscure route-quality failures: stock-terminated predictions may satisfy the formal benchmark objective while relying on chemically implausible transformations or artifacts inherited from patent-derived data~\cite{morgunov2025procrustean-retrocast}. We discuss additional limitations of \texttt{USPTO-190}, including low structural diversity and target relevance, in \autoref{app:uspto_190_critics}.

This limitation reflects a broader transition in the field from an era of navigability, where the central question was whether a planner could find any path through retrosynthetic search space, toward an era of validity, where the central question is whether that path is chemically plausible~\cite{Morgunov205Syntax}. Several recent benchmarks move closer to this validity-oriented setting. RetroTrim introduced 32 synthetically challenging, practically relevant targets without prior reported synthetic precedents~\cite{sadowski2025retrotrim-hallucinations}. This is an important step, but the set is too small for broad tool comparison and has not been released as an extensible benchmark family. URSA is designed to fill this gap by pairing practically relevant target sets with a unified route-level plausibility protocol, allowing both conventional CASP systems and LLM-based planners to be evaluated on the same chemical validity criteria.

%% file: text/main/preliminaries.tex
\section{Preliminaries}

\paragraph{Solv-N formalism}
The \texttt{Solv-N[task]} hierarchy separates increasingly strict notions of synthetic route validity~\cite{Morgunov205Syntax} from satisfaction of \texttt{[task]} constraints (e.g. termination to a stock of SMs): Solv-0 requires only basic syntactic validity, i.e. all molecular strings in the route to be valid SMILES. Solv-1 additionally requires each reaction step to correspond to a legal reaction template or otherwise valid reaction-center transformation. This requirement is usually explicit for template-based planners, where the reaction center is encoded in the template, but it is less guaranteed for template-free or generative outputs, which can hallucinate chemically invalid transformations~\cite{tetko2024models-matter, sadowski2025retrotrim-hallucinations}. In practice, chemists often assess questionable transformations by searching precedents in curated reaction databases such as SciFinder and Reaxys~\cite{scifinder, reaxys}, but route-level expert review is too slow and expensive to serve as a routine benchmark protocol~\cite{maziarz2025syntheseus}. Solv-2, the focus of this work, adds chemical plausibility constraints beyond template legality, including reaction precedent, chemo-, regio-, and stereoselectivity consistency. Solv-3 requires experimental executability, including conditions, purification, and yield. URSA focuses on Solv-2 because evaluating Solv-3 executability is only meaningful after the route has first passed stock termination and reaction-level chemical validity checks.

\paragraph{ChemCensor}
ChemCensor is a reaction-level plausibility evaluator based on synthetic precedent~\cite{zagribelnyy2026chemcensor}. It abstracts reported reactions into reaction centers and functional-group contexts, then uses these patterns to assess whether a proposed transformation is chemically plausible, including checks related to reaction-center precedent and chemo-, regio-, and stereoselectivity. In URSA, ChemCensor is used as a candidate automated validator for applying Solv-2 plausibility checks across every step of a proposed route. Because recent work has proposed LLMs as reaction or route evaluators~\cite{Schwaller2026-synthegy}, we benchmark ChemCensor against LLM-based judgments before using it as the route-level plausibility component of URSA.

%% file: text/main/approach.tex
\section{Utilitarian Retrosynthesis Assessment}
\label{sec:approach}

URSA (Utilitarian RetroSynthesis Assessment) is a route-level evaluation framework for testing whether synthesis planning systems produce chemically plausible stock-terminated routes. It is designed to compare heterogeneous planners, including conventional CASP systems and LLM-based approaches, under the same Solv-0, Solv-1, and Solv-2 criteria. In the URSA \texttt{[task]} configuration used here, routes at every reported Solv-N level are additionally required to terminate in the specified stock of commercially available building blocks (CABBs).

\subsection{Benchmarking sets}

URSA evaluates planners on two complementary target-molecule sets. The first, \texttt{URSA-expert-2026} (adopted from the ChemCensor study), contains novel, not-yet-synthesized target molecules whose synthetic accessibility was assessed by professional synthetic chemists through reference-supported route proposals. This set is more than three times larger than the novel target set proposed in RetroTrim~\cite{sadowski2025retrotrim-hallucinations}. To reduce leakage risk, the supporting expert-designed schemes are not used as route-reproduction ground truth in URSA; the benchmark evaluates predicted routes by chemical plausibility rather than by exact recovery of those schemes.

Because these expert targets are intentionally challenging, they may exceed the practical capabilities of many current synthesis planners. We therefore include a second, intermediate-difficulty benchmark, \texttt{URSA-drugs\&clinicals-2026}, consisting of 100 approved drugs and clinical candidates. This set is intended as a medicinally relevant replacement for \texttt{USPTO-190}: unlike patent-derived synthetic intermediates from \texttt{USPTO-full}~\cite{Lowe2017usptofull}, its targets are selected for practical relevance, structural diversity, and direct connection to real therapeutic chemistry. Together, the two sets allow URSA to evaluate both prospective planning on difficult novel molecules and route plausibility for known, clinically meaningful compounds. We additionally evaluate \texttt{URSA-drugs\&clinicals\&agrochemicals-2026}, a broader set comprising 300 drugs and clinical candidates and 33 agrochemicals which probes performance across a wider chemical space.

\begin{figure*}[t]
  \centering
  \includegraphics[
    %page=1,
    width=\linewidth,
    trim=50 120 50 100,
    clip
  ]{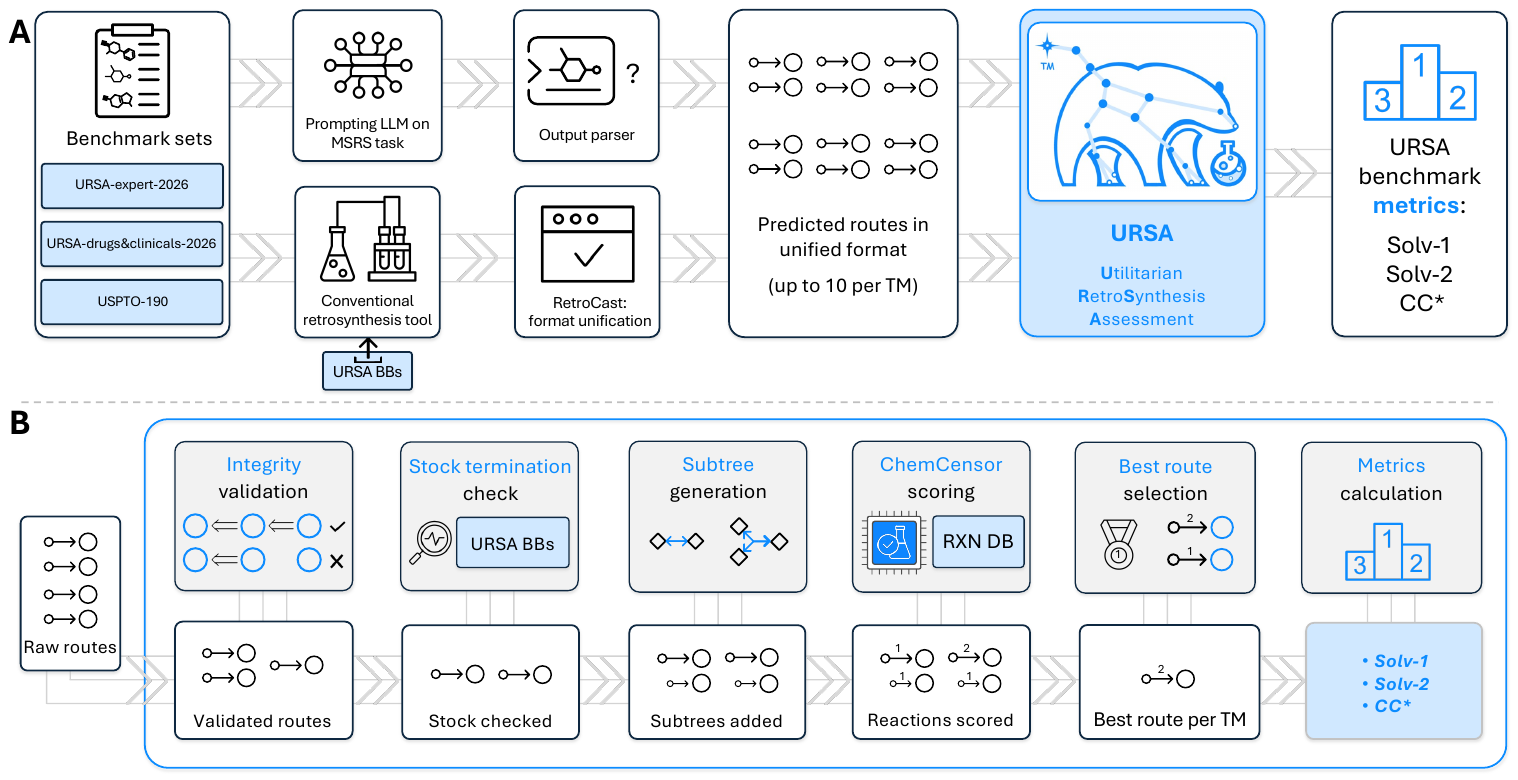}
  \caption{
URSA benchmark system. \textbf{A.} The benchmarking pipeline for both conventional synthesis planning tools and LLMs. \textbf{B.} The URSA benchmark components.
}
  \label{fig:CC_pipe}
\end{figure*}

\subsection{Evaluation protocol formalization}
\label{app:evaluation_protocol}

URSA evaluates the necessary precondition for executable synthesis planning under \texttt{Solv-N[task]} formalism: whether a proposed route is stock-terminated [task], structurally (Solv-0) and topologically (Solv-1) valid, and chemically plausible (Solv-2) at the level of its constituent transformations. A full Solv-3 evaluation would require conditions, catalysts, solvents, purification details, safety constraints, and expected yields. Those details are important for laboratory execution, but they are only meaningful after chemically invalid or selectively impossible transformations have already been filtered. URSA therefore focuses on Solv-2 rather than attempting premature condition-level evaluation.

For each target molecule, a submitted route must satisfy three requirements:

\begin{enumerate}
    \item The route must terminate in commercially available building blocks;
    \item The route must be structurally self-consistent, with valid molecular graphs and coherent directed-acyclic route connectivity;
    \item Each reaction in the route must be chemically plausible under the CC-based Solv-2 criterion.
\end{enumerate}

Solv-2 is conjunctive: failure of any requirement, including a single implausible reaction, assigns Solv-2 = 0 to the route. The mean ChemCensor score is used to rank candidate routes and contributes to the separately reported CC*, but cannot override a failed Solv-2 check (see \autoref{app:route_level_solv2}).

Because synthesis planning systems emit incompatible route formats, URSA first standardizes all predictions. Conventional CASP outputs are converted into the RetroCast route representation~\cite{morgunov2025procrustean-retrocast}; LLM-generated routes are parsed from the prompt-specified format described in \autoref{app:msrs_templates}. This produces a common route object for validation, scoring, visualization, and metric computation.

The pipeline then applies a sequence of route-level checks (\autoref{fig:CC_pipe}). First, primary integrity validation confirms that each route is a coherent directed acyclic graph, removes meaningless steps, and verifies that all molecular strings can be parsed by RDKit~\cite{landrum2023rdkit}. Second, stock-termination checking verifies that terminal leaves are commercially available building blocks. Third, URSA expands validated routes through subtree generation. This module enumerates shorter alternative subroutes by collapsing one-component reaction steps and merging adjacent reaction nodes when appropriate; already merged steps are not collapsed again. The procedure is illustrated in \autoref{app:subtree_gen_example}.

\begin{figure*}[th]
  \centering
  \includegraphics[
    %page=1,
    width=0.8\linewidth,
    trim=0 350 180 10,
    clip
  ]{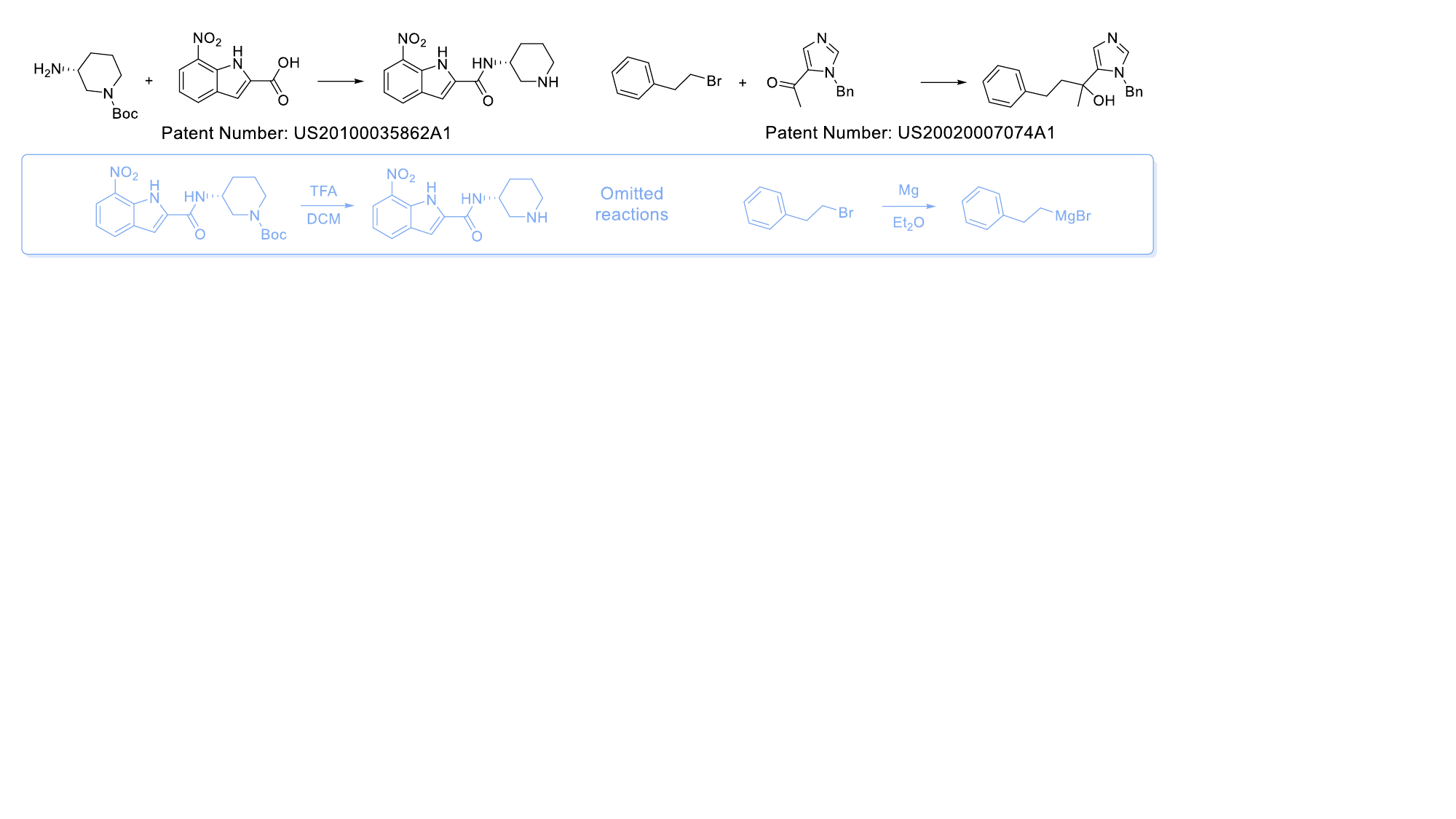}
  \caption{
Examples of multi-step reaction entities stored in the USPTO reaction database.
}
  \label{fig:multi_step_uspto}
\end{figure*}

Subtree generation addresses a practical mismatch between predicted routes and precedent databases. Reaction datasets such as USPTO often store multi-step operations as single reaction records, omitting intermediates for transformations such as Boc deprotection or Grignard reagent preparation (\autoref{fig:multi_step_uspto}). Without route augmentation, a predicted single-step transformation may be unfairly compared against a multi-step precedent, or vice versa. Subtree generation, therefore, creates alternative route views that reduce scoring artifacts caused by different levels of reaction granularity.

After validation and augmentation, URSA extracts every reaction step and scores it with ChemCensor using a reference database of synthetic precedents. ChemCensor (CC) scoring components are described in \autoref{app:chemcensor_rc}. Step-level scores are then aggregated into route-level scores. For each target molecule, URSA selects the best route if several ones are provided. There are two best route selection policies: CC-centric and length-centric. The results reported here use the first policy which implies the following priority order:

\begin{enumerate}
    \item Prefer routes with Solv-2 = 1, meaning every reaction passes the plausibility criterion;
    \item If multiple Solv-2 routes are available, select the route with the highest average CC score;
    \item If no route satisfies Solv-2, select the route with the highest average CC score;
    \item If multiple routes have the same average CC score, select the shortest route.
\end{enumerate}

In the length-centric policy, Solv-2 status remains the first priority, but among routes with the same status,
the shortest route is selected. The CC-centric policy favors routes with stronger ChemCensor support
across their steps, whereas the length-centric policy favors fewer transformations and the results of its application can be found in \autoref{app:route_selection_policy}.

URSA reports model performance using Solv-0, Solv-1, Solv-2, and the average ChemCensor plausibility score (CC*, \autoref{eq:chemcensor}). Best routes can also be exported as CDXML files annotated with ChemCensor scores for visualization and manual analysis.

%% file: text/main/experiments.tex
\section{Experimental Methods}
\label{sec:experiments}

This section describes the experimental design used to evaluate URSA. We run two linked experiments. First, we compare ChemCensor (\autoref{sec:code_and_data_availability}, \cite{zagribelnyy2026chemcensor}) with LLM-based reaction judges on an expert-labeled reaction plausibility dataset; this determines whether ChemCensor is suitable as the automated Solv-2 validator used by URSA. Second, using ChemCensor as the route-level plausibility component, we benchmark conventional CASP planners and LLM-based planners on the URSA target sets.

\subsection{Evaluated systems}
We evaluate LLMs, using their default reasoning-effort settings, in two roles: as reaction-level plausibility judges in the validator comparison and as route-generators in the route-level benchmark. ChemCensor is evaluated only as a reaction-level validator and then used as the Solv-2 scoring component in URSA. Conventional CASP tools are evaluated only as route-generating planners. The complete list of evaluated LLMs and conventional CASP tools is provided in \autoref{app:evaluated systems}.

\subsection{Reaction-level validator comparison}
\label{subsec:plausibility}
Following recent proposals to use LLMs for retrosynthesis evaluation~\cite{Schwaller2026-synthegy}, we compare the LLMs listed above with ChemCensor, a deterministic chemical-plausibility evaluator~\cite{zagribelnyy2026chemcensor}. The goal of this experiment is to select a route-level Solv-2 validator for URSA.

The reaction plausibility benchmark (\texttt{URSA-reaction-plausibility-bench-2026}) contains 1000 reactions collected from synthesis-planning model outputs by expert chemists. Each reaction was labeled as plausible (\textbf{1}) or implausible (\textbf{0}); the final highly balanced dataset contains 500 plausible and 500 implausible reactions (see \autoref{app:plausibility_benchmark_characterization}). LLMs were evaluated with a derivative of the prompt proposed in~\cite{Schwaller2026-synthegy}, provided in \autoref{app:plausibility_prompt}, and were required to output a binary plausibility label. ChemCensor scores on the \texttt{[0,1-5]} scale were binarized as \texttt{0: [0]} and \texttt{1: [1-5]}. Examples of LLM completions are provided in \autoref{app:plausibility_examples} and the accuracy and MCC (Matthews Correlation Coefficient) values for the best-performing models are reported in \autoref{tab:plausibility}, while the complete results for all evaluated models are provided in \autoref{app:plausibility_full_results}.

\subsection{Route-level planning benchmark}
\label{subsec:route_benchmark}
The second experiment evaluates synthesis planners under the URSA protocol, using ChemCensor as the Solv-2 scoring component for generated routes. All planners are evaluated against a shared starting-material stock containing 255,365 compounds (\autoref{sec:code_and_data_availability}). The stock combines Enamine building blocks~\cite{enamine}, the ASKCOS \texttt{buyables} set~\cite{askcos_2025}, and common small-molecule reagents or solvents that are often omitted from vendor stocks but are needed for realistic stock termination. The \texttt{buyables} component was additionally cross-checked against eMolecules ready-to-purchase Tiers $1$ and $2$ building blocks~\cite{emolecules}. Target molecules are drawn from three benchmark sets: \texttt{URSA-expert-2026}, \texttt{URSA-drugs\&clinicals-2026} (\autoref{sec:code_and_data_availability}), and the conventional \texttt{USPTO-190} set~\cite{chen2020retrolearningretrosyntheticplanning}.

\begin{table*}[tb!]
\centering
\input{tables/main/plausibility}
\caption{Reaction plausibility accuracy and MCC on 1,000 expert-labeled reactions.}
\label{tab:plausibility}
\end{table*}

LLM planners were evaluated with natural-language route-generation prompts; the full templates are provided in \autoref{app:msrs_templates}. Each model was allowed to provide up to 10 routes per target molecule. If a model returned more than 10 routes, outputs were truncated by the model-provided route scores when available, or randomly sampled when no scores were provided. For each target molecule, URSA selects a best route according to the protocol in \autoref{app:evaluation_protocol}. Final model performance is reported using Solv-0, Solv-1, Solv-2, and the average per-target ChemCensor score. The average ChemCensor score is computed over the best route for each target:

\begin{equation}
\mathrm{CC}^{*} = \frac{1}{M} \sum_{t=1}^{M} 
\underbrace{\frac{1}{\bigl|r^{*}(t)\bigr|} 
\sum_{i \,\in\, r^{*}(t)} \mathrm{CC}_{i}}_{\text{Av.\ PR ChemCensor for best route}}\label{eq:chemcensor}
\end{equation}

where $M$ is the number of target molecules, $r^{*}(t)$ is the best route selected for target $t$, $|r^{*}(t)|$ is the number of reactions in that route, and $\mathrm{CC}_{i} \in [0,5]$ is the ChemCensor score for reaction $i$. Unlike Solv-N[task], which reports the fraction of targets for which a stock-terminated route meets the corresponding validity criterion, CC* averages the mean ChemCensor step score of the selected route for each target on a 0–5 scale. Reaction steps can be scored even when their proposed route fails the stock constraint. So, a nonzero CC* does not imply Solv-0, Solv-1, or Solv-2 success. We therefore use Solv-2 as the primary measure of target-level success and CC* as a diagnostic of reaction-level chemical plausibility. Best routes are released as CDXML files with per-reaction ChemCensor annotations (\autoref{sec:code_and_data_availability}).

The primary benchmark configuration, \texttt{URSA-minor-1.3.0-U3} depends on ChemCensor v1.4.0 with the \texttt{U3} precedent database (\autoref{sec:code_and_data_availability}), containing approximately 1.4M reactions from \texttt{USPTO-full} prepared in ChemCensor-compatible format~\cite{Lowe2017usptofull}. We also report results for the closed \texttt{URSA-major-1.3.0-U3P4} configuration, which extends \texttt{U3} with the Pistachio 2026Q3 dataset~\cite{pistachio_nextmove}, adding approximately 3.8M synthetic precedents yielding \texttt{U3P4} database of synthetic precedents, and exploits the same ChemCensor v1.4.0 reaction-level plausibility engine.

Main route-level benchmark results are summarized in \autoref{tab:solvability_banchmark} and \autoref{fig:solv_values_by_model}. Additional \texttt{URSA-major-1.3.0} results, \texttt{USPTO-190} and \texttt{URSA-drugs\&clinicals\&agrochemicals} results, inference costs, and detailed analyses are provided in \autoref{app:solbabily_benchmark_appendix}.

%% file: tables/main/plausibility.tex
\begin{tabular}{lccccc}
\toprule
\textbf{Model} & ChemCensor v1.4.0 & Grok-4.5 & Gemini 3.1 Pro & GPT 5.5 & Claude Opus 4.8 \\
\midrule
\textbf{Accuracy} & \textbf{0.96} & 0.80 & 0.83 & 0.76 & 0.73 \\
\textbf{MCC}      & \textbf{0.92} & 0.62 & 0.67 & 0.57 & 0.52 \\
\bottomrule
\end{tabular}

%% file: text/main/results.tex
\section{Main Findings and Discussion}
\label{sec:main_results}

\subsection{ChemCensor outperforms LLM judges on reaction plausibility}

ChemCensor achieves the highest accuracy on the expert-labeled reaction plausibility benchmark (0.96), while requiring only local execution and no API inference cost (\autoref{tab:plausibility}). The best LLM judge, Gemini 3.1 Pro, reaches only 0.83 accuracy and only 0.67 MCC. GPT 5.5, Grok-4.5, and Claude Opus 4.8  cluster
between 0.73 and 0.80 of the accuracy values with the corresponding MCC values in the 0.52-0.62 range. 

Although the LLM judges show nontrivial discrimination ability, their accuracy substantially overstates their balanced classification performance. All evaluated LLMs exhibit a systematic bias toward labeling reactions as plausible: recall for plausible reactions is high (0.90–0.96), whereas recall for non-plausible reactions is markedly lower (0.50–0.73). This asymmetry leads to frequent false-positive plausibility assignments and explains the considerably lower MCC values despite superficially reasonable accuracy. In contrast, ChemCensor maintains strong discrimination between both classes. From a route-planning perspective, tolerance for false positives is particularly detrimental in the early stages of retrosynthesis, as it can lead to a combinatorial proliferation of implausible branches in the retrosynthetic search tree. These results and the critical assessment of the Synthegy approach (see \autoref{app:synthefy_critics}) support using deterministic ChemCensor as the automated Solv-2 validator in URSA, rather than relying on direct LLM judgments of reaction plausibility.

\subsection{Stock termination remains far above chemical validity}

\begin{figure}[b!]
  \centering
  \includegraphics[trim=0 0 0 30, clip, width=0.8\linewidth]{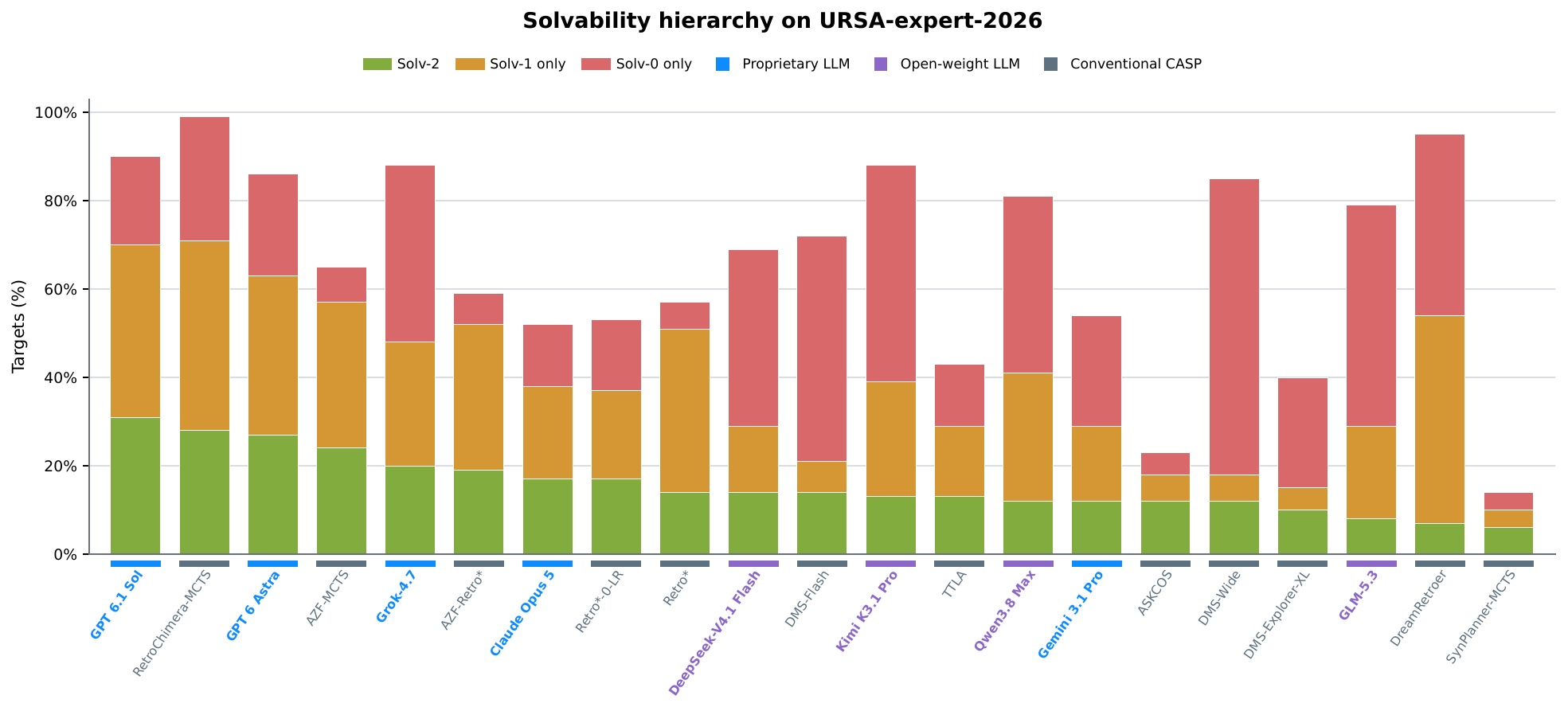}
  \caption{Solv-N hierarchy on \texttt{URSA-expert-2026}, with models ranked by Solv-2. Each bar sums to Solv-0 and comprises \colorbox{ursaGreen}{Solv-2}, \colorbox{ursaAmber}{Solv-1 only}, and \colorbox{ursaCoral}{Solv-0 only}.}
  \label{fig:solv_values_by_model}
\end{figure}

The route-level benchmark shows that each stricter Solv-N level filters out a substantial fraction of generated routes ( \autoref{fig:solv_values_by_model}, \autoref{tab:solvability_banchmark}). This confirms the central premise of URSA: stock termination is necessary as a key task constraint to be satisfied, but it is not sufficient for chemically meaningful synthesis planning.

On \texttt{URSA-expert-2026}, GPT 6.1 Sol achieves the highest Solv-2 score (31\%), followed by RetroChimera (28\%) and GPT 6 Astra (27\%). Even the leading system solves fewer than one-third of these intentionally challenging targets. The gap between stock termination and chemical plausibility is substantial: RetroChimera reaches 99\% Solv-0 but 28\% Solv-2, while DreamRetroer drops from 95\% to 7\%. So, high stock-termination rates do not ensure chemically plausible routes.

\begin{table}[t!]
\centering
\input{tables/main/solvability}
\caption{Solvability benchmark. \textbf{CC*} is the average ChemCensor per step along the best route, averaged across targets (see \autoref{eq:chemcensor}). Benchmark version: \texttt{URSA-minor-1.3.0-U3}, ChemCensor v.1.4.0, \texttt{U3} DB of synthetic precedents. See \autoref{tab:solvability_banchmark_full} for the complete results.}
\label{tab:solvability_banchmark}
\end{table}

Performance is higher on \texttt{URSA-drugs\&clinicals-2026}, consistent with this set being more accessible and closer to familiar medicinal chemistry. The best Solv-2 values are achieved by RetroChimera-MCTS (54\%), AZF-MCTS and AZF-Retro* (51\%), followed by GPT 6 Astra (46\%). In this setting, conventional planners remain highly competitive with frontier LLMs and often outperform them. The stronger results on this set indicate that URSA can distinguish between difficult prospective-style targets and known clinically meaningful molecules without collapsing both into a single stock-termination score.

To assess the sensitivity of model comparisons to the target set, we evaluated eight selected systems on the expanded, 333-target \texttt{URSA-drugs\&clinicals\&agrochemicals-2026} benchmark. Across the eight systems evaluated on both benchmark sets, Solv-2 decreases by 7.8 pp on average on the 333-target extension (see \autoref{app:dca_results}). Nevertheless, the Solv-2 rankings remain strongly concordant (Spearman's $\rho = 0.98$): 26 of 27 non-tied model pairs retain their order, and the three leading systems remain unchanged. These results suggest that the primary 100-target set captures much of the comparative signal with approximately 70\% fewer target evaluations, while the extension probes a broader chemical space.

\subsection{Synthesis planning across LLM generations}

\begin{figure}[t]
  \centering
  \includegraphics[trim=5 5 5 20, clip, width=0.90\linewidth]{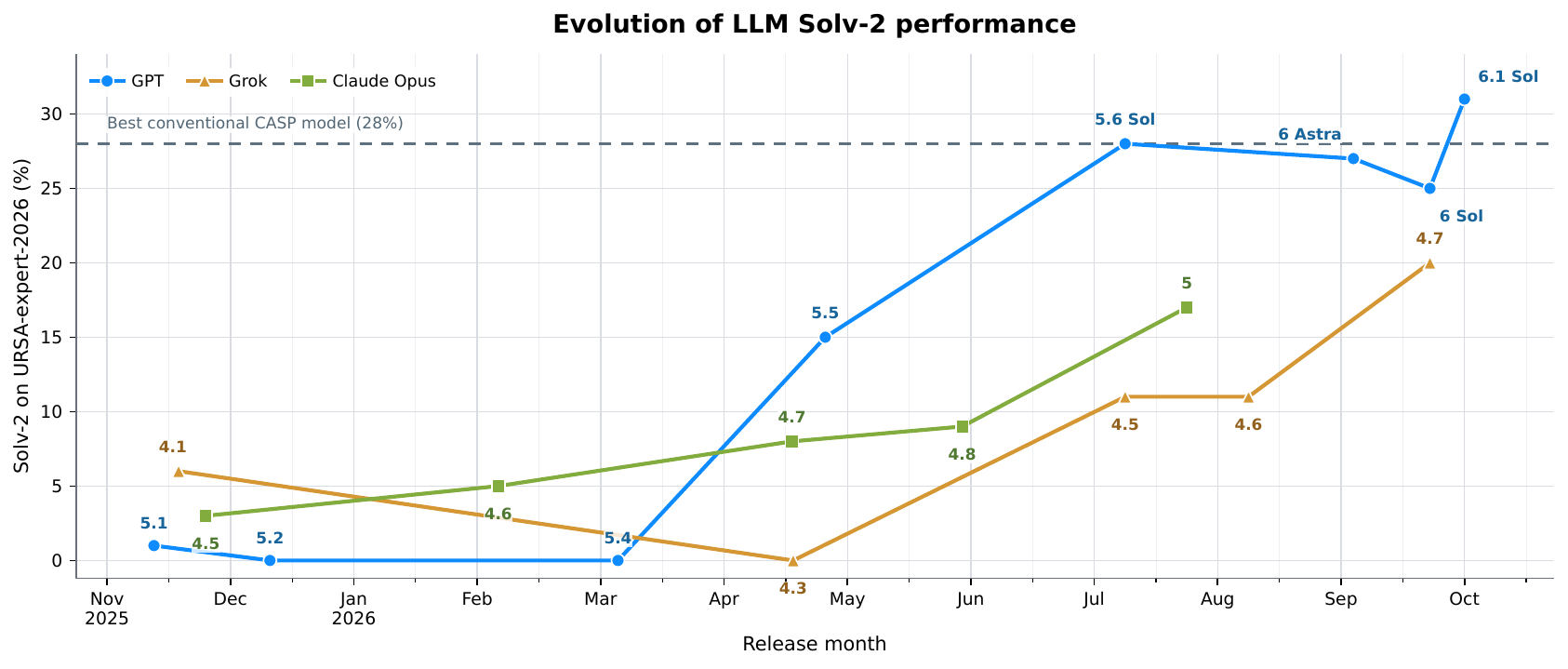}
  \caption{Evolution of LLM performance on multi-step retrosynthesis task.}
  \label{fig:solv2_timeline}
\end{figure}

Spanning a significant period relative to the fast-moving LLM landscape, this paper captures the evolution of retrosynthetic applicability across LLM generations. The progress of three families of general-purpose models studied in the frame of this work is outlined in \autoref{fig:solv2_timeline}. We can see that the GPT models start outperforming the rivals starting from the GPT 5.5 release. Expectedly, the reasons lying behind synthesis planning failures are noticeably changing when comparing old and new versions (\autoref{app:failure_decompose}). The stock termination rate as well as the valid route syntax are the main hurdles in the multi-step retrosynthesis task for earlier models. As LLMs progress, they gain some understanding of typical building blocks and struggle less with achieving the Solv-0 criterion. The current most advanced GPT models are able to predict routes which not only converge to the URSA stock but also pass Solv-1 and Solv-2 criteria with absolute metrics higher than those obtained for classical CASP tools. The gap between Solv-1 and Solv-2 values underpins that functional group compatibility remains a challenge for some transformations and leaves room for model improvement.

\subsection{Practical considerations}

\begin{figure}[b]
  \centering
  \includegraphics[trim=7 5 0 30, clip, width=0.9\linewidth]{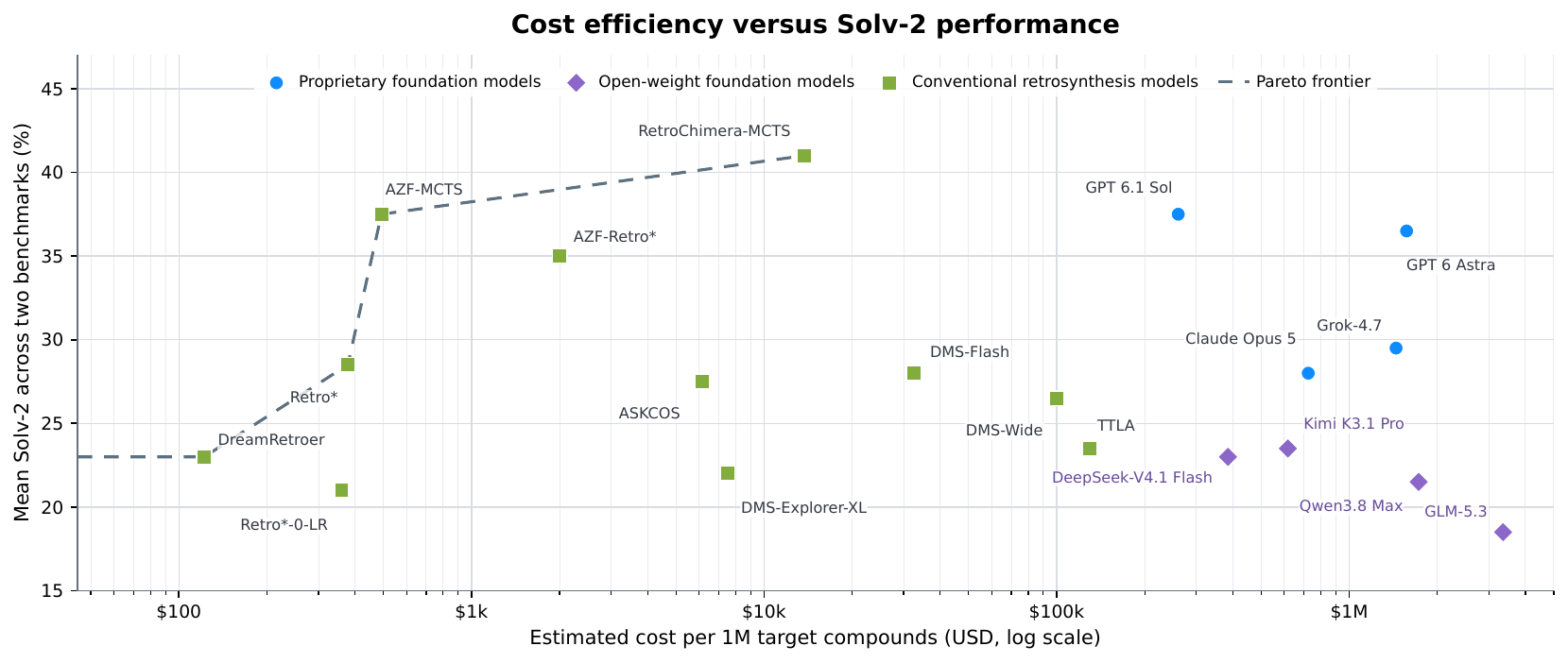}
  \caption{Cost versus mean Solv-2 performance.}
  \label{fig:pareto}
\end{figure}

Cost further differentiates the practical utility of the benchmarked systems (see \autoref{fig:pareto}). Among LLMs with recorded token usage, GPT 6.1 Sol achieves the highest Solv-2 on \texttt{URSA-expert-2026} (31\%) at an estimated API cost of \$27.60 for the 100-target run, while GPT 6 Astra leads the LLMs on \texttt{URSA-drugs\&clinicals-2026} (46\%) at \$166.22 (\autoref{tab:costs}). Conventional planners achieve competitive or higher Solv-2 at much lower estimated run costs: AZF-MCTS reaches 24\% and 51\% for \$0.05 and \$0.04, and RetroChimera-MCTS reaches 28\% and 54\% for \$1.43 and \$1.31 on the two sets, respectively (\autoref{tab:costs_conv}). RetroChimera-MCTS thus leads on the drug/clinical targets and falls only three percentage points short of the expert-set leader at approximately 19-fold lower cost. These estimates compare LLM API charges with conventional-planner instance costs, but make the cost--validity trade-off visible beyond Solv-N rates.

Expanding the ChemCensor precedent database from the open \texttt{URSA-minor-1.3.0-U3} configuration to the closed \texttt{URSA-major-1.3.0-U3P4} configuration increases absolute Solv-2 values for many systems, but does not substantially change the overall ranking (\autoref{app:u3p4}). This result suggests that precedent coverage materially affects plausibility scoring and that larger open reaction corpora would improve the reliability of automated Solv-2 benchmarks that derive chemical-plausibility judgments from synthetic precedents.

An important observation is the apparent decoupling between reaction plausibility assessment and multistep synthesis planning capability. While frontier LLMs demonstrate relatively similar performance on the isolated reaction plausibility discrimination task (\autoref{tab:plausibility}), their performance diverges substantially under route-level Solv-2 evaluation (\autoref{tab:solvability_banchmark}). This suggests that the principal challenge of retrosynthesis may no longer lie in recognizing locally plausible transformations, but rather in maintaining globally coherent synthetic trajectories under long-horizon chemical constraints. The result further supports the view that synthesis planning cannot be reduced to independent single-step reaction evaluation.

%% file: tables/main/solvability.tex
\begin{tabular}{@{}l|llll|llll@{}}
\toprule
\multirow{2}{*}{\textbf{Model}}
& \multicolumn{4}{c|}{\textit{URSA-expert-2026}}
& \multicolumn{4}{c}{\textit{URSA-drugs\&clinicals-2026}} \\
\cmidrule(lr){2-5}\cmidrule(lr){6-9}
& \textbf{Solv-0} & \textbf{Solv-1} & \textbf{Solv-2} & \textbf{CC*}
& \textbf{Solv-0} & \textbf{Solv-1} & \textbf{Solv-2} & \textbf{CC*} \\
\midrule
\multicolumn{9}{c}{\textit{Proprietary Foundation Models}} \\
\midrule
Grok-4.7                & 88 & 48 & 20 & 1.58 & 92 & 60 & 39 & 2.33 \\
Gemini 3.1 Pro          & 54 & 29 & 12 & 1.45 & 70 & 47 & 38 & 2.42 \\
GPT 6 Astra             & 86 & 63 & \cellcolor{bronze!40}27 & \textbf{1.94} & 85 & 67 & 46 & 2.44 \\
GPT 6.1 Sol             & 90 & \underline{70} & \cellcolor{gold!40}31 & 1.87 & 87 & 70 & 44 & 2.40 \\
Claude Opus 5           & 52 & 38 & 17 & 1.66 & 70 & 53 & 39 & \textbf{2.58} \\
\midrule
\multicolumn{9}{c}{\textit{Open-weight Foundation Models}} \\
\midrule
DeepSeek-V4.1 Flash     & 69 & 29 & 14 & 1.35 & 78 & 47 & 32 & 2.23 \\
Qwen3.8 Max             & 81 & 41 & 12 & 1.67 & 79 & 49 & 31 & 2.37 \\
Kimi K3.1 Pro           & 88 & 39 & 13 & 1.46 & 92 & 55 & 34 & 2.23 \\
GLM-5.3                 & 79 & 29 & 8 & 1.44 & 89 & 47 & 29 & 2.08 \\
\midrule
\multicolumn{9}{c}{\textit{Conventional Retrosynthesis Models}} \\
\midrule
Retro*                  & 57 & 51 & 14 & 1.36 & 76 & 64 & 43 & 2.35 \\
AZF-MCTS                & 65 & 57 & 24 & 1.58 & 81 & \underline{72} & \underline{51} & 2.26 \\
AZF-Retro*              & 59 & 52 & 19 & 1.40 & 74 & 70 & \underline{51} & 2.31 \\
DMS-Flash               & 72 & 21 & 14 & \underline{1.89} & 78 & 52 & 42 & \underline{2.53} \\
Retro*-0-LR             & 53 & 37 & 17 & 1.36 & 58 & 38 & 25 & 1.85 \\
SynPlanner-MCTS         & 14 & 10 & 6 & 1.84 & 38 & 37 & 28 & 2.43 \\
ASKCOS                  & 23 & 18 & 12 & 1.39 & 60 & 50 & 43 & 2.39 \\
RetroChimera-MCTS       & \textbf{99} & \textbf{71} & \cellcolor{silver!40}28 & 1.49 & \textbf{100} & \textbf{73} & \textbf{54} & 2.22 \\
DreamRetroer            & \underline{95} & 54 & 7 & 0.95 & \underline{97} & 62 & 39 & 1.44 \\
TTLA                    & 43 & 29 & 13 & 1.34 & 53 & 45 & 34 & 2.29 \\
\bottomrule
\end{tabular}

%% file: text/main/limitations.tex
\section{Limitations}
\label{sec:limitations}
URSA currently evaluates route validity through Solv-2, not Solv-3 executability: conditions, catalysts, solvents, purification strategies, safety constraints, and expected yields remain outside the scoring framework. Its Solv-2 judgments also depend on the coverage and biases of the ChemCensor precedent database, which in the open configuration is primarily patent-derived. Finally, the main metrics are computed on the selected best route per target molecule, so they do not fully characterize route diversity, ranking quality, or failure distributions across all generated candidates.

%% file: text/main/conclusion.tex
\section{Conclusion}
\label{sec:conclusion}
URSA reframes retrosynthesis benchmarking around chemical validity rather than the ability to reach purchasable starting materials alone. In the terminology of the \texttt{Solv-N[task]} framework, URSA is intended as a benchmark for the field's transition from navigability-oriented evaluation to validity-oriented evaluation. Across conventional CASP systems and LLM-based planners, stock-terminated routes often fail stricter Solv-1 and Solv-2 checks, showing that apparent solvability can substantially overestimate practical synthesis capability. By pairing standardized route parsing with ChemCensor-based plausibility scoring, URSA enables scalable route-level evaluation that would be impractical with manual expert review. 

The results further show that multistep synthesis planning is not reducible to isolated reaction plausibility assessment. Strong local reaction judgment does not guarantee chemically coherent trajectories across an entire route, especially on difficult prospective targets. Just as ChemCensor provides a one-to-many alternative to single-ground-truth Top-K accuracy in single-step retrosynthesis, URSA-based Solv-N benchmarking offers a way to move beyond both Top-K route accuracy and formal solvability in multistep retrosynthesis evaluation. We will continue evaluating newly released CASP systems and LLM-based planners with URSA and publish updated benchmark results on \url{https://dddbench.insilico.com/}. Future progress will require planners that combine route-level generation with stronger chemical constraints, broader open precedent corpora, and eventually Solv-3 modules for executable laboratory detail.

%% file: text/main/availability.tex
\section{Code and Data Availability}
\label{sec:code_and_data_availability}
The live URSA leaderboard is available at \url{https://dddbench.insilico.com/}. All datasets in this work are released on Hugging Face: the benchmark sets at \url{https://huggingface.co/datasets/insilicomedicine/URSA-benchmarking-sets} and the building block set at \url{https://huggingface.co/datasets/insilicomedicine/URSA-BBs}. Result artifacts (best routes as annotated CDXML files) are archived on Zenodo at \url{https://doi.org/10.5281/zenodo.20810716}. Per-artifact details are provided in \autoref{app:data_availability}.

The source code is available on GitHub:
\begin{itemize}
    \item URSA framework: \url{https://github.com/insilicomedicine/URSA};
    \item ChemCensor: \url{https://github.com/insilicomedicine/ChemCensor}.
\end{itemize}

The ChemCensor repository provides the protocol and scripts for constructing its reference database. The URSA repository provides the full LLM benchmarking workflow, from prompting to route evaluation and metric computation.

%% file: text/main/funding.tex
\section*{Acknowledgments and Disclosure of Funding}

We thank our colleagues at Insilico Medicine who assisted with benchmarking the evaluated models, as well as the anonymous NeurIPS reviewers for their constructive feedback.

This work was supported by Insilico Medicine.
All authors except A.M. are employees of Insilico Medicine, a group of commercial AI companies. A.M. contributed as an independent researcher.

%% file: text/main/impact.tex
\section{Impact Statement}
\label{app:impact_statement}

Rational synthesis planning is a daily challenge for millions of organic chemists worldwide, from graduate students designing their first multi-step routes to medicinal chemists in pharmaceutical companies racing to bring new therapeutics to patients. The ability to objectively evaluate whether a proposed synthetic route is chemically sound, rather than only whether it terminates in purchasable starting materials, has the potential to accelerate drug discovery, reduce chemical waste, and lower the barrier to high-quality retrosynthesis for researchers without deep synthetic expertise. By providing a rigorous, chemistry-aware benchmarking framework, URSA aims to raise the standard by which synthesis planning systems are compared and ultimately improved, and we hope it contributes meaningfully to that goal.

At the same time, we must acknowledge an inherent dual-use tension. Improved benchmarks drive development of more capable synthesis planning systems, and the same chemical reasoning that guides the synthesis of a life-saving drug applies equally to the preparation of toxic or otherwise harmful compounds. Organic chemistry does not discriminate between benign and malicious intent. URSA and ChemCensor evaluate chemical plausibility against a corpus of reported transformations; they do not and cannot filter by the nature of the target molecule. More capable planners, validated by more rigorous benchmarks, may therefore also lower the barrier to the synthesis of dangerous substances. We believe this risk is best managed through community norms, institutional oversight, and responsible disclosure practices rather than through artificial restrictions on scientific tools. And we encourage the community to develop access policies and safeguards for synthesis planning systems proportional to their capabilities.

%% file: text/appendix/Code_and_data_availablity.tex
\section{Data Availability Details}
\label{app:data_availability}

\begin{itemize}
    \item \textit{Benchmark sets} (\texttt{URSA-reaction-plausibility-bench-2026}, \texttt{URSA-expert-2026}, \texttt{URSA-drugs\&clinicals-2026} and \texttt{URSA-drugs\&clinicals\&agrochemicals-2026}) --- provided as separate CSV files at \url{https://huggingface.co/datasets/insilicomedicine/URSA-benchmarking-sets}.

    \item \textit{URSA Building Blocks} --- the starting-material stock, provided as a CSV file at \url{https://huggingface.co/datasets/insilicomedicine/URSA-BBs}.

    \item \textit{Best routes from URSA} --- the single best route per target molecule for every benchmarking set, exported as CDXML files with per-reaction ChemCensor (CC) scores. Files are organized into six folders by benchmarking set and tool category (conventional synthesis planner or LLM), with one subfolder per model. Archived on Zenodo at \url{https://doi.org/10.5281/zenodo.20810716}.
\end{itemize}

%% file: text/appendix/glossary.tex
\section{Glossary}
\label{app:gloassary}

\begin{description}

\item[\textbf{Target molecule}]
The target molecule is the desired chemical compound that represents the ultimate goal of synthesis planning. It is the molecule for which the system generates or evaluates synthetic routes, and it serves as the starting point for retrosynthetic disconnection, working backward from the target to identify precursor molecules.

\item[\textbf{Reactants}]
Reactants are the chemical compounds that undergo transformation during a chemical reaction and whose atoms are directly incorporated into the product structure. Technically, reactants are distinguished from reagents by the presence of atom mapping -- atoms in reactants have corresponding mapped atoms in the product(s).

\item[\textbf{Reagents}]
Reagents are chemical compounds that participate in a chemical reaction but whose atoms are not directly incorporated into the product structure. Reagents typically facilitate or enable the transformation (e.g., catalysts, bases, acids, solvents with reactive roles) with no or only minor atom contribution to the final product. For benchmarking, it is reasonable to extend the reacting species with reagents, since they may influence the correctness of atom--atom mapping and thereby the reaction-center extraction process.

\item[\textbf{Starting materials}]
Starting materials are the initial chemical compounds from which a synthetic route begins. They are molecules that exist at the terminal nodes (leaves) of a retrosynthetic tree and are not produced by any reaction step within the route. Starting materials serve as the input chemicals for the synthesis and are expected to be commercially available or otherwise accessible within the reported synthetic methods.

\item[\textbf{Building blocks}]
Building blocks (CABBs) are commercially available chemical compounds that can be found in vendor datasets and purchased from them.

\item[\textbf{Single-step retrosynthesis model}]
An SSRS model predicts one or several retrosynthetic disconnections by mapping a target product molecule to a set of precursor reactants corresponding to a single reaction step. The model does not perform recursive planning or multi-step route construction, focusing instead on identifying chemically plausible reactants.

\item[\textbf{Multi-step retrosynthesis model}]
An MSRS model is a system that applies retrosynthetic transformations (typically recursively) to decompose a target molecule into commercially available or otherwise accessible starting materials through a sequence of reaction steps.

\item[\textbf{Reaction center (RC)}]
An RC is the set of atoms (dynamic atoms) in one or more reactant molecules and the product molecule that undergoes change during a chemical transformation, including atoms/bonds that are formed, broken, created, destroyed, or whose connectivity, bond order, formal charge, or hybridization state differs between reactants and products.

\item[\textbf{Reaction center component}]
One single-site transformation. Technically, it's obtained by splitting a full extracted reaction center along atom-map overlap between reactant and product fragments: parts that share maps are grouped so each group is one independent simultaneous transformation (e.g. two protections cleaved in one step yields two components). Reaction centers which contain several components are named composite or multi-component.

\item[\textbf{Chemical plausibility}]
Chemical plausibility reflects alignment of a reaction with core principles of organic synthesis (e.g., chemoselectivity, regioselectivity, stereoselectivity). Operationally, it can be reduced to chemoinformatic concepts such as reaction centers, functional groups, their occurrence, and compatibility rules, potentially augmented with conditions (solvents, temperature, catalysts, auxiliary reagents). In this system, plausibility is assessed by comparing the reaction center and functional-group context against a reference dataset of verified transformations. If the extracted reaction center is absent from the reference library, the reaction is considered implausible; similarly, functional groups never observed for that reaction center negate plausibility. When both reaction-center and functional-group context are supported by precedents, the reaction is considered chemically plausible.

\item[\textbf{Level of confidence}]
The nominal degree of chemical plausibility is estimated via discrete levels of confidence (LC), depending on which reaction-center representation is matched among verified transformations. Higher LC indicates that a more specific (larger-context) reaction-center definition is supported, correlating with higher nominal plausibility and representativeness in terms of synthetic precedents. The LC value is used as the reaction score in ChemCensor.

\item[\textbf{Functional groups (FGs)}]
FGs are structural motifs that determine chemical reactivity and properties. In the present system, functional groups are represented as SMARTS patterns \cite{smarts} that can be matched to molecular structures via substructure search. Functional-group context annotated for each reaction center helps determine which patterns are tolerable for a transformation, supporting chemical plausibility.

\item[\textbf{Functional group (FG) signature}]
An FG signature is the ensemble of FGs present in reactant/product molecules that are not affected by the transformation. For a given reaction center, the signature is constructed by aggregating synthetic precedents from the reference dataset.

\item[\textbf{ChemCensor}]
ChemCensor Score (CC Score) is a quantitative metric evaluating the chemical plausibility of a reaction and/or a retrosynthetic route by measuring the proportion of steps that pass plausibility validation, weighted by their confidence levels.

\item[\textbf{Solvability}]
Solvability is a formal metric indicating either the presence (1) or absence (0) of at least one synthetic route for a target molecule to starting materials, or the fraction (percentage) of targets in a set for which a complete route is proposed.

\item[\textbf{Solv-0}]
(Syntactic Solvability) The lowest tier of the validity hierarchy, requiring only that all molecules in a proposed route are syntactically valid graphs (i.e., obey rules of valency and aromaticity).

\item[\textbf{Solv-1}]
(Topological Solvability) The second tier of the validity hierarchy, requiring that a route successfully connects molecules via topologically valid reaction steps. In the means of ChemCensor-based Solv-1 calculation, Solv-1 requires all reactions from a route to have synthetic precedents for their reaction centers. 

\item[\textbf{Solv-2}]
(Selectivity Solvability) The third tier of the validity hierarchy, requiring that each transformation in a route is chemically plausible by satisfying constraints of chemoselectivity, regioselectivity, diastereoselectivity, enantioselectivity, and stoichiometry.

\item[\textbf{Solv-3}]
(Executability Solvability) The highest tier of the validity hierarchy, requiring that a route is experimentally viable under realistic laboratory conditions, accounting for factors such as yield, purification, safety, cost, and reagent availability

\item[\textbf{Reference dataset}]
A reference dataset is a collection of verified reaction transformations extracted from sources including patents (e.g., USPTO), articles, preprints, and ELNs. It may include metadata such as conditions and yield. In this system, the reference dataset is used to validate analyzed reactions via reaction-center matching and functional-group signature comparison.

\item[\textbf{Synthetic precedent}]
A synthetic precedent is an elementary synthetic fact of a successful chemical reaction recorded in a reference dataset.

\item[\textbf{Distributivity of FG signature}]
A property of a multi-component center which implies that FGs of individual reaction center components when found in the reference dataset are added to FG signature of this complex center. Only FGs common for all components are considered.

\end{description}

%% file: text/appendix/evaluated_systems.tex
\section{Evaluated systems}
\label{app:evaluated systems}

\textbf{Proprietary Foundation Models:} Grok 4.1 \cite{xai2025grok41}, 4.3 \cite{xai2026grok43}, 4.5 \cite{xai2026grok45}, 4.6 \cite{xai2026grok46}, 4.7 \cite{xai2026grok47}; Gemini 3.1 Pro \cite{google2026gemini3.1pro};  GPT 5.1 \cite{openai2025gpt51}, 5.2 \cite{openai2025gpt52}, 5.4 \cite{openai2025gpt54}, 5.5 \cite{openai2025gpt55} 5.6 family (Luna, Terra, Sol) \cite{openai2025gpt56}, 6 Astra \cite{openai2025gpt6astra}, 6 Luna and Sol \cite{openai2025gpt6sol_luna}, 6.1 Sol \cite{openai2025gpt6_1_sol}; Claude Sonnet 4.5 \cite{anthropic2025claudesonnet45} and 4.6 \cite{anthropic2025claudesonnet46}; Claude Opus 4.5 \cite{anthropic2025claudeopus45}, 4.6 \cite{anthropic2025claudeopus46}, 4.7 \cite{anthropic2025claudeopus47}, 4.8 \cite{anthropic2026claudeopus48}, 5 \cite{anthropic2026claudeopus5}. This list lacks Gemini 4 Argon and Opus 5.5 due to, respectively, very low API throughput and too strict guardrails. \textbf{Open-weight Foundation Models:} DeepSeek-V3.2 \cite{deepseekai2025deepseekv32} and V4.1 Flash \cite{deepseekai2025deepseekv4.1}; Qwen3.5-397B-A17B (hereafter Qwen3.5) \cite{QwenTeam2026qwen35} and 3.8-Max \cite{QwenTeam2026qwen38}; Kimi K2.5 \cite{kimiteam2026kimik25visualagentic}, K3 \cite{kimiteam2026kimik3openfrontier} and K3.1 Pro \cite{kimiteam2026kimik3.1}; GLM-5 \cite{glm5team2026glm5vibecodingagentic} and 5.3 \cite{glm5team2026glm53}. The reasoning efforts for the models are provided in \autoref{tab:solvability_banchmark_full} and the same levels of reasoning were used in all other tasks across the study.

\textbf{Conventional CASP tools:} The evaluated conventional synthesis planning tools were either ready-to-use end-to-end solutions, such as Retro* \cite{chen2020retrolearningretrosyntheticplanning}; DirectMultiStep (DMS) family of models \cite{shee2025directmultistep}: DMS-Wide, DMS-Flash and DMS-Explorer-XL; ASKCOS \cite{askcos_2025}; TTLA \cite{Kreutter2023ttla}; DreamRetroer \cite{dreamretroer_2025}, SynLlama \cite{synllama_2025} or composite solution of (\textit{i}) single-step retrosynthesis policy LocalRetro (LR) \cite{Chen2021localretro} and (\textit{ii}) orchestrating planning algorithm Retro*-0 implemented via Syntheseus \cite{maziarz2025syntheseus}. AiZynthFinder \cite{aizyn_2020, Saigiridharan2024AiZynthFinder} was benchmarked in two modes: with Retro* and MCTS as planners and max route depth of 10. RetroChimera \cite{retrochimera_2025} was benchmarked via the Syntheseus interface \cite{retrochimera_github} with MCTS as planner. SynPlanner \cite{synplanner_2024} was evaluated with the MCTS planner and max route depth of 10. The versions of these CASP tools are outlined in \autoref{tab:casp_versions}.

\begin{center}
\centering
\input{tables/appendix/casp_tools_versions}
\captionof{table}{Classical CASP tools and their versions.}
\label{tab:casp_versions}
\end{center}

%% file: tables/appendix/casp_tools_versions.tex
\label{tab:casp-versions}
\begin{tabular}{llp{7.5cm}}
\toprule
\textbf{Model} & \textbf{Version} & \textbf{Configuration} \\
\midrule
Retro*              & via Syntheseus 0.8.0 & default\\
DMS                 & 1.1.3            & Wide, Flash and Explorer-XL methods \\
Retro*-0-LR         & via Syntheseus 0.8.0 & default\\
ASKCOS              & 2.0             & default \\
DreamRetroer        & 1.0              & default\\
SynLlama            & 0.1.0            & default\\
TTLA                & 1.6.9            & default\\
SynPlanner-MCTS     & 1.3.2            & 100 iterations, depth 10 \\
AiZynthFinder (AZF) & 4.4.1            & depth 10, Retro* and MCTS planners \\
RetroChimera-MCTS   & 1.3.0            & Pistachio checkpoint\\
\bottomrule
\end{tabular}

%% file: text/appendix/plausibility_bencmark_set.tex
\section{URSA Reaction Plausibility Benchmark}
\label{app:reaction_plausibility_section}

\subsection{\texttt{URSA-reaction-plausibility-bench-2026} dataset}
\label{app:plausibility_benchmark_characterization}
\texttt{URSA-reaction-plausibility-bench-2026} contains 1,000 reactions collected from synthesis-planning model outputs and annotated by expert chemists: 500 plausible and 500 implausible. The released dataset provides reaction SMILES, the expert binary label, the ChemCensor score, and its binarized value. Expert labels serve as the reference for the validator comparison in \autoref{subsec:plausibility}; ChemCensor scores are included to make that comparison reproducible. The dataset is available at \url{https://huggingface.co/datasets/insilicomedicine/URSA-benchmarking-sets}.

\paragraph{Functional-group coverage.}
We matched reactant and product structures against the 465 patterns applicable to this analysis from the 516-pattern ChemCensor functional-group (FG) library. We applied the same procedure to approximately 56,000 reactions from PaRoutes~\cite{paroutes_2022}. A set of 95 FG types accounted for 90\% of all matched FG occurrences in the URSA benchmark and 91\% in PaRoutes. For pairwise combinations, 241 FG pairs accounted for 73\% and 72\% of all observed pairs, respectively. Thus, the two sets have similar cumulative coverage of common FG contexts despite their different sizes. This comparison concerns FG occurrences, not the fraction of reactions containing at least one matched FG.

\paragraph{Selectivity and stereochemistry.}
Chemoselectivity was identified as the principal problem in 224 of the 500 reactions labeled implausible by experts. Across the benchmark, 269 reactions test preservation of specified stereochemistry, while 57 test the emergence of stereochemical information in the product: 35 introduce an R/S configuration and 22 define double-bond geometry. A further 95 reactions contain a second chemically similar reactive center, such as an S\textsubscript{N}Ar transformation in the presence of another aryl halide. The set also includes 111 S\textsubscript{E}Ar reactions with competing C--H sites. These annotations describe potentially overlapping challenges and should not be interpreted as disjoint reaction classes.

\paragraph{Reaction classes.}
We classified reactions using Rxn-INSIGHT~\cite{dobbelaere2024rxninsight}, adapting the classification used by Schneider et al.~\cite{schneider2016bigdata}. We divided C--C bond formation into cross-coupling, olefination, organometallic reactions involving Grignard or organolithium reagents, and other transformations such as enolate alkylation, aldol condensation, and Friedel--Crafts reactions. The same classification was applied to an 8,000-reaction sample from PaRoutes. The distributions are reported in \autoref{tab:plausibility_reaction_classes}.

\begin{table}[h]
\centering
\small
\begin{tabular}{@{}p{0.55\linewidth}rr@{}}
\toprule
Reaction class & URSA (\%) & PaRoutes (\%) \\
\midrule
Heteroatom alkylation and arylation & 20.6 & 24.0 \\
\textbf{C--C bond formation (total)} & \textbf{29.1} & \textbf{12.5} \\
\quad Cross-coupling & 8.0 & 6.2 \\
\quad Olefination & 3.0 & 1.2 \\
\quad Organometallic (RMgX/RLi) & 5.7 & 1.5 \\
\quad Other C--C bond formation & 12.4 & 3.6 \\
Functional-group interconversion & 11.3 & 8.5 \\
Acylation & 10.8 & 18.8 \\
Heterocycle formation & 8.4 & 5.2 \\
Reduction & 5.7 & 10.7 \\
Deprotection & 5.4 & 13.4 \\
Functional-group addition & 5.7 & 4.0 \\
Protection & 1.2 & 1.7 \\
Oxidation & 1.0 & 1.4 \\
Resolution & 0.1 & 0.0 \\
\bottomrule
\end{tabular}
\caption{Reaction-class distribution in the 1,000-reaction \texttt{URSA-reaction-plausibility-bench-2026} benchmark set and an 8,000-reaction sample from PaRoutes. The C--C total is the sum of its four subclasses; percentages are rounded.}
\label{tab:plausibility_reaction_classes}
\end{table}

The benchmark contains the major reaction classes found in PaRoutes but assigns a larger share to C--C bond formation (29.1\% versus 12.5\%), particularly organometallic and other C--C transformations. The similar FG coverage therefore does not imply identical reaction-class distributions.

\subsection{Plausibility Prompt}
\label{app:plausibility_prompt}

\begin{tcolorbox}[
    title=System prompt,
    breakable,
    enhanced,
    fonttitle=\bfseries,
    colback=white,
    colframe=black!75,
]
\footnotesize

You are an expert organic chemist tasked with analyzing proposed chemical reactions and determining their plausibility.

Your goal is to provide a reaction plausibility assessment in a binary mode based on its reactants, products, mechanism, and other relevant factors. You will be asked to perform the analysis of a chemical reaction proposed by a retrosynthesis model. You must evaluate objectively and highlight any issues with the proposal.

The following reaction has not been experimentally validated. It only provides the desired pathway, however we want to assess the plausibility of this reaction as this is currently entirely theoretical.

Examine the reaction SMILES provided above carefully.

Using your expertise in organic chemistry, perform the following analysis:

\begin{enumerate}
    \item Identify the reactants and products in the reaction.
    \item Analyze the reaction mechanism, including:
    \begin{itemize}
        \item Bond formation and breaking
        \item Electron movement
        \item Intermediates (if any)
    \end{itemize}
\end{enumerate}

Identify key structural changes that occur during the reaction. Evaluate the reaction's characteristics:
\begin{itemize}
    \item Efficiency (yield, number of steps)
    \item Selectivity
    \item Reagents and conditions required
    \item Potential side products
\end{itemize}

Assess the reaction's plausibility, considering:
\begin{itemize}
    \item Electronics
    \item Sterics
    \item Mechanistic explanation
    \item Possible alternative pathways
    \item Selectivity of the proposed reaction
\end{itemize}

Before providing your final assessment, wrap your analysis in \texttt{<analysis>} tags. In your analysis:

\begin{enumerate}
    \item List the identified reactants and products separately.
    \item Reaction SMILES vs.\ heavy-atom accounting: confirm that every reactant or reagent whose heavy atoms are incorporated into the product is shown explicitly in the reaction SMILES (with species separated as usual, e.g.\ by \texttt{.} on the reactant side). Do not treat as a defect the omission of optional agents (e.g.\ catalysts, additives, solvents) whose heavy atoms never appear in the product---their absence is acceptable. Likewise, do not penalize entries that explicitly list such species when that is chemically reasonable; including them is optional and not a negative.
    \item Identify and list functional groups present in reactants and products.
    \item Describe the key steps in the reaction mechanism, including:
    \begin{itemize}
        \item Initial bond breaking events
        \item Formation of any intermediates
        \item Final bond formation events
    \end{itemize}
    \item Highlight the main structural changes that occur.
    \item Discuss the electronic and steric factors that influence the reaction's plausibility.
    \item Explain the mechanistic rationale for the proposed reaction.
    \item Consider and describe any potential alternative pathways or competing reactions.
\end{enumerate}

This structured approach will help ensure a thorough interpretation of the reaction and its citability.

After your analysis, provide a detailed assessment of the reaction in the following format:

\begin{tcolorbox}[
    colback=gray!8,
    colframe=gray!40,
    boxrule=0.4pt,
    left=6pt, right=6pt,
    fontupper=\ttfamily\small,
]
\footnotesize
\textbf{<assessment>}\\
Reaction Components: [List identified reactants and products]\\
Functional Groups: [List functional groups present in reactants and products]\\
Mechanism Overview: [Describe the key steps in the reaction mechanism]\\
Structural Changes: [Highlight the main structural changes that occur]\\
Electronic Factors: [Discuss how electronic effects influence the reaction's plausibility]\\
Steric Considerations: [Explain any steric factors that impact the reaction's plausibility]\\
Alternative Pathways: [Describe any potential alternative pathways or competing reactions]\\
Plausibility Analysis: [Provide a detailed assessment of the reaction's plausibility, including efficiency, selectivity, and mechanistic rationale]\\
Justification: [Explain your reasoning for the assigned binary label, referencing specific aspects of the reaction mechanism and characteristics]\\
\textbf{</assessment>}
\end{tcolorbox}

Finally, assign exactly one binary plausibility label:

\begin{itemize}
    \item \texttt{0} = not plausible (the reaction as written is not a plausible chemical transformation under reasonable conditions)
    \item \texttt{1} = plausible (the reaction is a plausible chemical transformation)
\end{itemize}

Present the label in the following format only: \texttt{<score>0</score>} or \texttt{<score>1</score>}

Remember to base your assessment solely on the information provided in the reaction SMILES and your expert knowledge of organic chemistry. Do not make assumptions about the reaction's intended use or refer to external data sources.

\end{tcolorbox}

\begin{tcolorbox}[
    title=Query Prompt,
    enhanced,
    fonttitle=\bfseries,
    colback=white,
    colframe=black!75,
]
\footnotesize
Reaction SMILES: \texttt{<\{\{input\_format\}\}>\textcolor{blue}{\{\{reaction\_smiles\}\}}</\{\{input\_format\}\}>}.
Give a binary plausibility label only: \texttt{0} = reaction is not plausible, \texttt{1} = reaction is plausible.
Briefly justify, then output exactly \texttt{<score>0</score>} or \texttt{<score>1</score>}.
\end{tcolorbox}

\subsection{Models Completion Examples}
\label{app:plausibility_examples}

\begin{windowbox}[width=\linewidth]{Model Input}
    \footnotesize
    \textbf{System Prompt:} \\
    <...> \\
    \textbf{Query Prompt:} \\
    Reaction SMILES: \texttt{<\{\{reaction\_smiles\}\}>\textcolor{blue}{CN.CNC[C@@H](O)c1ccc(C2(C=O)CCCC2)cc1>>\\CNC[C@@H](O)c1ccc(C2(CNC)CCCC2)cc1}</\{\{reaction\_smiles\}\}>}.
Give a binary plausibility label only: \texttt{0} = reaction is not plausible, \texttt{1} = reaction is plausible.
Briefly justify, then output exactly \texttt{<score>0</score>} or \texttt{<score>1</score>}.
\end{windowbox}

\begin{windowbox}[width=\linewidth]{Annotated Input}
\begin{fieldbox}{Reaction SMILES Visualization}
  \centering
  \includegraphics[width=0.7\linewidth, trim=0 460pt 530pt 0, clip]{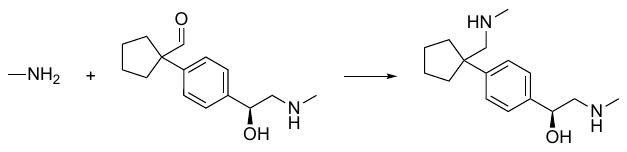}
  \label{fig:reaction_smiles}
\end{fieldbox}
\begin{fieldbox}{Expert Label}
\footnotesize
0
\end{fieldbox}
\begin{fieldbox}{Expert Comment}
\footnotesize
Potential regioselectivity issue -- two amino-groups
\end{fieldbox}
\end{windowbox}

\vspace{0.8em}

\begin{windowbox}[width=\linewidth]{Model: Gemini 3.1 Pro}
\footnotesize
<analysis>
Reaction Components:
Reactants: 
1. Methylamine (CN)
2. 1-(4-((1R)-1-hydroxy-2-(methylamino)ethyl)phenyl)cyclopentane-1-carbaldehyde (CNC[C@@H](O)c1ccc(C2(C=O)CCCC2)cc1)

Product: 
1. (1R)-2-(methylamino)-1-(4-(1-((methylamino)methyl)cyclopentyl)phenyl)ethan-1-ol (CNC[C@@H](O)c1ccc(C2(CNC)CCCC2)cc1)

Functional Groups:
Reactants: Primary amine (methylamine), secondary amine, secondary alcohol, aldehyde, aromatic ring, cyclopentane ring.
Product: Secondary amines (two), secondary alcohol, aromatic ring, cyclopentane ring.

Mechanism Overview:
The reaction is a reductive amination. 
1. The primary amine (methylamine) acts as a nucleophile and attacks the electrophilic carbonyl carbon of the aldehyde.
2. This forms a hemiaminal intermediate, which subsequently undergoes dehydration (loss of a water molecule) to form an imine (-CH=N-CH3).
3. The imine intermediate is then reduced by a reducing agent (which is conventionally omitted in the reaction SMILES, such as NaBH(OAc)3, NaBH3CN, or H2/Pd) to yield the final secondary amine product (-CH2-NH-CH3).

Structural Changes:
The aldehyde group (-CHO) attached to the quaternary carbon of the cyclopentane ring is converted into a methylaminomethyl group (-CH2NHCH3). The rest of the molecule, including the chiral center, remains unchanged.

Electronic Factors:
Aldehydes are highly electrophilic and readily undergo nucleophilic attack by primary amines. The presence of the electron-rich aromatic ring does not negatively impact the electrophilicity of the aldehyde since they are separated by a quaternary sp3 carbon, preventing any direct conjugation.

Steric Considerations:
The aldehyde is situated on a quaternary carbon (a neopentyl-like position), which introduces some steric hindrance. However, aldehydes are generally reactive enough that imine formation and subsequent reduction can still proceed efficiently, especially with a small, unhindered nucleophile like methylamine.

Alternative Pathways:
The starting material contains a secondary amine which could theoretically react with the aldehyde intermolecularly to form an iminium ion, leading to dimerization after reduction. However, methylamine is a smaller, less sterically hindered primary amine and is typically used in large excess in such reactions. This kinetic and concentration advantage ensures that the reaction with methylamine outcompetes any potential dimerization.

Plausibility Analysis:
The proposed transformation is a standard reductive amination. The heavy-atom accounting is perfectly consistent, with the aldehyde carbon and the methylamine combining to form the new -CH2-NH-CH3 group. The loss of the aldehyde oxygen as water is a standard feature of this mechanism. The omission of the reducing agent in the reaction SMILES is acceptable and standard practice for such cheminformatics representations. The reaction is highly plausible under standard reductive amination conditions (e.g., using NaBH(OAc)3 in dichloroethane or methanol).

Justification:
The reaction is a textbook reductive amination of an aldehyde using methylamine. The structural changes are chemically sound, the heavy atoms are properly accounted for, and the potential side reactions can be easily mitigated by standard reaction conditions (e.g., excess methylamine). The steric hindrance from the quaternary center is not severe enough to prevent the reaction. Therefore, the reaction is highly plausible.
</analysis>

<score>1</score>
\end{windowbox}

\vspace{0.8em}

\begin{windowbox}[width=\linewidth]{Model: GPT 5.5}
\footnotesize
Reductive amination of a formyl substituent with methylamine to give the N-methylaminomethyl product is chemically plausible, assuming a suitable reducing agent/conditions are used. Competing imine formation or reductive amination at the aldehyde is reasonable.

<score>1</score>
\end{windowbox}

\newpage
\subsection{Reaction Plausibility Benchmark Complete Results}
\label{app:plausibility_full_results}

\begin{table}[h]
\centering
\begin{tabular}{@{}llcc@{}}
\toprule
\textbf{Model} & \textbf{Reasoning effort} & \textbf{Accuracy} & \textbf{MCC} \\
\midrule
GPT 5.5             & medium       & 0.76 & 0.57 \\
GPT 5.6 Luna        & high         & 0.59 & 0.31 \\
GPT 5.6 Terra       & high         & 0.60 & 0.32 \\
GPT 5.6 Sol         & high         & 0.63 & 0.38 \\
GPT 6 Luna          & high         & 0.67 & 0.40 \\
GPT 6 Sol           & high         & 0.71 & 0.51 \\
GPT 6 Astra         & medium       & 0.71 & 0.52 \\
GPT 6.1 Sol         & medium       & 0.71 & 0.50 \\
Gemini 3.1 Pro      & high         & 0.83 & 0.67 \\
Grok 4.5            & high         & 0.80 & 0.62 \\
Grok 4.6            & high         & 0.73 & 0.51 \\
Grok 4.7            & high         & 0.75 & 0.52 \\
Claude Opus 4.8     & high         & 0.73 & 0.52 \\
Claude Opus 5       & high         & 0.71 & 0.51 \\
Kimi K3             & max          & 0.65 & 0.39 \\
Kimi K3.1 Pro       & high         & 0.65 & 0.41 \\
DeepSeek V4.1 Flash & high         & 0.65 & 0.39 \\
Qwen 3.8 Max        & xhigh        & 0.66 & 0.41 \\
\bottomrule
\end{tabular}
\caption{Reaction plausibility benchmark results for all evaluated LLMs.}
\label{tab:plausibility_all_models}
\end{table}

%% file: text/appendix/URSA-expert-2026.tex
\section{\expertset{} Set} 
\label{app:expert_set_app}

The 2D structures of molecules from the URSA-expert-2026 set are provided in \cite{zagribelnyy2026chemcensor}.

The dataset diversity is $0.85886$. Dataset diversity is defined as the average pairwise dissimilarity between all distinct molecule pairs:
\begin{equation}
\mathrm{Diversity} = \frac{1}{N(N-1)} \sum_{i \ne j} \left( 1 - \mathrm{Similarity}(i,j) \right),
\label{eq:diversity}
\end{equation}
where $N$ is the number of molecules in the dataset.

Molecular similarity in \autoref{eq:diversity} is computed as the cosine similarity between binary vectors of structural screens, where each vector element indicates the presence or absence of a specific atom-centered structural fragment derived using the Chemosoft software (ChemDiv Inc. chemical database software, \url{https://www.chemdiv.com/}). The set of screens is constructed dynamically across the dataset, capturing atom types, bond types, and ring environments without using hash-based compression.

%% file: text/appendix/Sovability_benchmark.tex
\clearpage
\section{URSA Benchmark. Additional Experiments.}
\label{app:solbabily_benchmark_appendix}

\subsection{Complete benchmark results for all evaluated models}

\begin{table}[th!]
\centering
\small
\input{tables/appendix/solvability_full}
\caption{Solvability benchmark. \textbf{CC*} is the average ChemCensor per step along the best route, averaged across targets (see \autoref{eq:chemcensor}). Benchmark version: \texttt{URSA-minor-1.3.0-U3}, ChemCensor v.1.4.0, \texttt{U3} DB of synthetic precedents.}
\label{tab:solvability_banchmark_full}
\end{table}

\clearpage
\subsection{URSA-major-U3P4}
\label{app:u3p4}

Table~\ref{tab:solvability_banchmark_u2p2} presents the \texttt{URSA-major-1.3.0-U3P4} (based on relational DB extended with Pistachio 2026Q3 \cite{pistachio_nextmove} dataset) benchmark results.

\begin{table*}[th!]
\centering
\small
\input{tables/appendix/u2p2}
\caption{Solvability benchmark results. \textbf{CC*} is the average ChemCensor per step along the best route, averaged across targets (see \autoref{eq:chemcensor}). Benchmark version: \texttt{URSA-major-1.3.0-U3P4}, ChemCensor v.1.4.0, \texttt{U3P4} DB of synthetic precedents.}
\label{tab:solvability_banchmark_u2p2}
\end{table*}

\clearpage
\subsection{URSA benchmark on USPTO-190 set}

Table~\ref{tab:uspto_results} presents the benchmarking results on the USPTO-190 dataset evaluated under the \texttt{URSA-minor-1.3.0-U3} and \texttt{URSA-major-1.3.0-U3P4} settings.

% \begin{center}
\begin{table}[th!]
\centering
% \small
\input{tables/appendix/uspto-190}
\caption{USPTO-190 solvability benchmark results on \texttt{URSA-minor-1.3.0-U3} and \texttt{URSA-major-1.3.0-U3P4}; ChemCensor v.1.4.0.}
\label{tab:uspto_results}
\end{table}

\subsection{Inference Costs and Resource Usage}
\label{sec:model_inferences}

We estimated LLM inference costs from recorded input and output token counts using the per-token rates in \autoref{tab:providers}. The rates are listed in USD per million tokens; reasoning tokens are included in the output-token counts. The estimates do not apply cache, batch, or off-peak discounts.

\begin{table}[H]
\centering
\input{tables/appendix/providers}
\caption{Pricing assumptions used to estimate LLM inference costs.}
\label{tab:providers}
\end{table}

\autoref{tab:costs} reports token usage and estimated cost for multistep route generation on the two URSA target sets. The route-level runs use 10 queries per target for each 100-target set; the reaction-level run evaluates 1,000 reactions. Gemini 3.1 Pro has benchmark results, but its token-usage records were unavailable for this cost analysis. Provider-reported token counts are not strictly comparable across APIs
because providers differ in tokenization and in their treatment of
cached and server-injected prompt tokens.

\begin{table}[H]
\centering
\input{tables/appendix/costs}
\caption{LLM token usage and estimated inference costs. $N_{\mathrm{in}}$ and $N_{\mathrm{out}}$ are in thousands of tokens; $C$ is in USD.}
\label{tab:costs}
\end{table}

For conventional retrosynthesis models, costs were estimated from observed wall-clock runtime and the hourly price of the compute instance. The resulting estimates for the two URSA target sets are reported in \autoref{tab:costs_conv}.

\begin{table}[H]
\centering
\input{tables/appendix/costs_conv}
\caption{Runtime and estimated cost of conventional synthesis planners. $T$ is wall-clock time in minutes, $H$ is instance price in USD per hour, and $R$ is estimated run cost in USD.}
\label{tab:costs_conv}
\end{table}

\subsection{Failure Decomposition for different versions of GPT models} 
\label{app:failure_decompose}

\begin{figure}[h!]
  \centering
  \includegraphics[trim=3 5 15 5, clip, width=\linewidth]{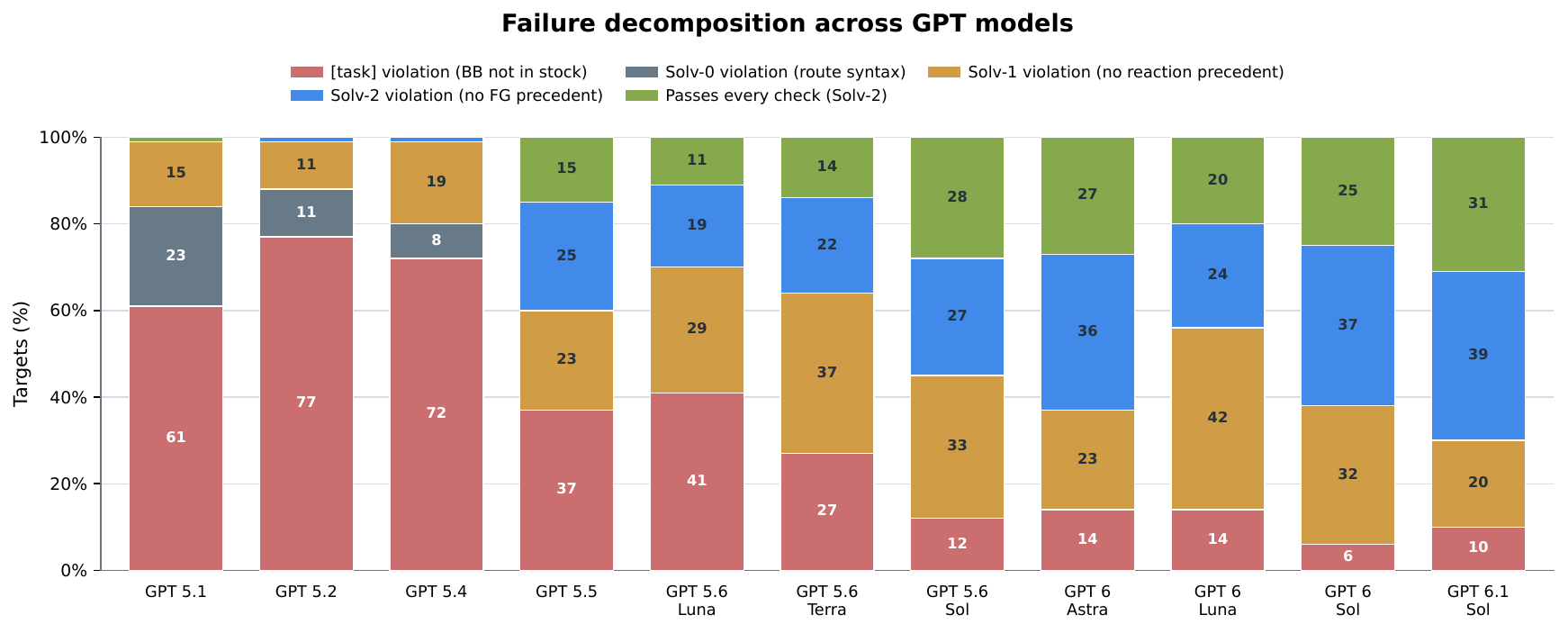}
  \caption{Evolution of GPT model performance on the multi-step retrosynthesis task.}
  \label{fig:gpt_family}
\end{figure}

\autoref{fig:gpt_family} shows how GPT models evolve by decomposing different violations of the \texttt{Solv-N[task]} criteria on \texttt{URSA-expert-2026}. The newer the version, the less the model struggles with basic route validity and the stock termination criterion.

\subsection{Benchmarking Statistics}

For each LLM, we issued 10 queries per target molecule, randomly sampling a prompt template
at each query (see \autoref{app:msrs_templates}). Although the primary metrics are computed from the single best route per target,
we employed non-parametric bootstrap resampling ($B = 10{,}000$ iterations) over targets to
estimate 95\% confidence intervals for the main metrics. The bootstrap 95\% confidence intervals for LLMs \textit{Solv-2} on the \texttt{URSA-expert-2026} are reported in \autoref{tab:bootstrap_ci}.

\begin{table}[H]
\centering
\begin{tabular}{@{}l|cc@{}}
\toprule
\textbf{Model} & \textbf{Solv-2} & \textbf{95\% CI} \\
\midrule
\multicolumn{3}{c}{\textit{Proprietary Foundation Models}} \\
\midrule
Grok-4.1               & 0.06 & [0.02, 0.11] \\
Grok-4.3               & 0.00 & [0.00, 0.00] \\
Grok-4.5               & 0.11 & [0.05, 0.18] \\
Grok-4.6               & 0.11 & [0.05, 0.17] \\
Grok-4.7               & 0.20 & [0.13, 0.28] \\
Gemini 3.1 Pro         & 0.12 & [0.06, 0.19] \\
GPT 5.1                & 0.01 & [0.00, 0.03] \\
GPT 5.2                & 0.00 & [0.00, 0.00] \\
GPT 5.4                & 0.00 & [0.00, 0.00] \\
GPT 5.5                & 0.15 & [0.08, 0.22] \\
GPT 5.6 Luna           & 0.11 & [0.05, 0.17] \\
GPT 5.6 Terra          & 0.14 & [0.08, 0.21] \\
GPT 5.6 Sol            & 0.28 & [0.19, 0.37] \\
GPT 6 Astra            & 0.27 & [0.19, 0.36] \\
GPT 6 Luna             & 0.20 & [0.13, 0.28] \\
GPT 6 Sol              & 0.25 & [0.17, 0.34] \\
GPT 6.1 Sol            & 0.31 & [0.22, 0.40] \\
Claude Sonnet 4.5      & 0.01 & [0.00, 0.03] \\
Claude Sonnet 4.6      & 0.03 & [0.00, 0.07] \\
Claude Opus 4.5        & 0.03 & [0.00, 0.07] \\
Claude Opus 4.6        & 0.05 & [0.01, 0.10] \\
Claude Opus 4.7        & 0.08 & [0.03, 0.13] \\
Claude Opus 4.8        & 0.09 & [0.04, 0.15] \\
Claude Opus 5          & 0.17 & [0.10, 0.24] \\
\midrule
\multicolumn{3}{c}{\textit{Open-weight Foundation Models}} \\
\midrule
DeepSeek-V3.2            & 0.00 & [0.00, 0.00] \\
DeepSeek-V4.1 Flash      & 0.14 & [0.07, 0.21] \\
Qwen3.5-397B-A17B        & 0.02 & [0.00, 0.05] \\
Qwen3.8 Max              & 0.12 & [0.06, 0.19] \\
Kimi K2.5                & 0.04 & [0.01, 0.08] \\
Kimi K3 Max              & 0.08 & [0.03, 0.14] \\
Kimi K3.1 Pro            & 0.13 & [0.07, 0.20] \\
GLM-5                    & 0.02 & [0.00, 0.05] \\
GLM-5.3                  & 0.08 & [0.03, 0.14] \\
\bottomrule
\end{tabular}
\caption{Bootstrap 95\% confidence intervals for Solv-2 ($B = 10{,}000$) for LLMs for \texttt{URSA-expert-2026} dataset.}
\label{tab:bootstrap_ci}
\end{table}

\subsection{URSA Evaluation Runtime}
\label{app:ursa_runtime}

We measured the end-to-end runtime of URSA~1.3.0 with
ChemCensor~1.4.0 on the 100-target \texttt{URSA-expert-2026}
benchmark, retaining at most 10 routes per target. The measurements used
the U3 ChemCensor database and included input parsing, route validation,
building-block checking, subtree generation, ChemCensor initialization
and scoring, best-route selection, and metric aggregation.
The experiments were performed on an AWS instance with one AMD EPYC
7R32 processor (32 physical cores, 64 hardware threads) and 249\,GiB of
RAM. Each run was restricted to 16 logical CPUs and used 16 ChemCensor
worker processes. GPU execution was explicitly disabled. OpenMP and
BLAS libraries were restricted to one thread per process to avoid
oversubscription. Each model was evaluated in a separate process using
a single wall-clock run.

\begin{table}[t]
\centering
\small
\setlength{\tabcolsep}{4pt}
\begin{tabular}{@{}lrrr|rr|rr@{}}
\toprule
\textbf{Model}
& \shortstack{\textbf{Targets}\\\textbf{with route}}
& \textbf{Routes}
& \shortstack{\textbf{Reactions}\\\textbf{scored}}
& \multicolumn{2}{c|}{\textbf{U3}}
& \multicolumn{2}{c}{\textbf{U3P4}} \\
\cmidrule(lr){5-6}\cmidrule(lr){7-8}
& & & &
\textbf{Total (s)} & \textbf{s/target} &
\textbf{Total (s)} & \textbf{s/target} \\
\midrule
AZF--MCTS       & 100 & 781 & 2963 &  85.6 & 0.86 &  99.0 & 0.99 \\
Gemini 3.1 Pro  & 100 & 896 & 1265 &  76.1 & 0.76 &  90.9 & 0.91 \\
GPT 5.5         & 100 & 955 & 2264 &  82.6 & 0.83 &  97.4 & 0.97 \\
Claude Opus 4.7 & 100 & 881 &  896 &  69.9 & 0.70 &  84.2 & 0.84 \\
RetroChimera    &  99 & 969 & 3285 & 101.5 & 1.02 & 115.2 & 1.15 \\
\bottomrule
\end{tabular}
\caption{End-to-end CPU runtime on \texttt{URSA-expert-2026}.
Time per target uses all 100 benchmark targets, including targets
without a valid proposed route.}
\label{tab:ursa_runtime}
\end{table}

ChemCensor and building-block-catalog initialization required
48.6--49.2\,s per run. ChemCensor scoring required 15.2--35.5\,s,
depending on the number and complexity of the unique reactions.

\subsection{Sensitivity to Planner-Provided Route Ranking}
\label{app:ranking_sensitivity}

The primary URSA protocol selects the best route after evaluating
the submitted candidates. To assess sensitivity to the planner's
original ordering, we repeat this selection using only the first
$k \in \{1,3,5,10\}$ routes for each target. Solv-2@$k$ is the
percentage of targets for which at least one of these routes passes
the route-level Solv-2 criterion. CC*@$k$ applies the usual
per-target aggregation to the route selected from the first $k$
candidates. For rank-weighted CC*@$k$, the selected route's mean
ChemCensor score is additionally multiplied by $j^{-1/2}$ before
aggregation, where $j$ is its original one-based planner rank.
We evaluate four planners that produced at least five routes per
target on average.

\begin{table}[h]
\centering
\renewcommand{\arraystretch}{1.05}
\setlength{\tabcolsep}{6pt}
\begin{tabular}{@{}lrrrrrr@{}}
\toprule
\textbf{Model}
& \textbf{Avg. routes}
& \boldmath{$k$}
& \textbf{Solv-1}
& \textbf{Solv-2}
& \textbf{CC*}
& \textbf{wCC*} \\
\midrule
\multirow{4}{*}{Retro*-0-LR}
& \multirow{4}{*}{9.5}
& 1  & 33 &  9 & 1.14 & 1.14 \\
& & 3  & 35 & 14 & 1.25 & 1.12 \\
& & 5  & 36 & 16 & 1.28 & 1.03 \\
& & 10 & 37 & 17 & 1.36 & 0.92 \\
\addlinespace
\multirow{4}{*}{AZF--MCTS}
& \multirow{4}{*}{7.8}
& 1  & 50 & 15 & 1.31 & 1.31 \\
& & 3  & 55 & 21 & 1.44 & 1.17 \\
& & 5  & 55 & 23 & 1.53 & 1.12 \\
& & 10 & 57 & 24 & 1.58 & 1.05 \\
\addlinespace
\multirow{4}{*}{ASKCOS}
& \multirow{4}{*}{8.9}
& 1  & 16 &  5 & 1.16 & 1.16 \\
& & 3  & 18 &  8 & 1.29 & 1.05 \\
& & 5  & 18 & 10 & 1.37 & 1.00 \\
& & 10 & 18 & 12 & 1.39 & 0.95 \\
\addlinespace
\multirow{4}{*}{TTLA}
& \multirow{4}{*}{9.5}
& 1  & 20 &  6 & 0.89 & 0.89 \\
& & 3  & 23 &  8 & 1.08 & 0.91 \\
& & 5  & 27 & 12 & 1.24 & 0.88 \\
& & 10 & 29 & 13 & 1.34 & 0.82 \\
\bottomrule
\end{tabular}
\caption{Sensitivity to planner-provided route ranking on
\texttt{URSA-expert-2026}, evaluated using URSA~1.3.0,
ChemCensor~1.4.0, and the U3 database. Solv-1 and Solv-2 are
percentages of all 100 targets.}
\label{tab:ranking_sensitivity}
\end{table}

Best-route selection measures whether a planner generates at least one Solv-2-valid route among its candidates, whereas practical use also depends on the planner-provided route order. For the four planners producing at least five routes per target on average, Solv-2@1 ranges from 5\% to 15\% and Solv-2@10 from 12\% to 24\% on \texttt{URSA-expert-2026}. All four gain 7--9 percentage points as $k$ increases from 1 to 10: AZF-MCTS rises from 15\% to 24\%, and Retro*-0-LR from 9\% to 17\%. Their relative order does not reverse, but the lower top-1 values show the practical importance of planner-provided ranking.

\subsection{Sensitivity to best-route selection policy}
\label{app:route_selection_policy}

We compared the CC-centric policy used in the main benchmark with a length-centric policy on the same candidate routes. Both policies prioritize Solv-2 status; among routes with the same status, the former selects the route with the highest mean ChemCensor score, whereas the latter selects the shortest route. Target-level Solv-2 is therefore unchanged by the policy choice, although the selected route and its CC* may differ. Switching to the length-centric policy reduced the mean selected-route length from 3.6 to 2.9 reaction steps with U3 scoring and from 3.7 to 2.7 steps with U3P4 scoring. This comparison shows the effect of the secondary selection criterion on the representative routes reported by URSA.

%% file: tables/appendix/solvability_full.tex
\begin{tabular}{@{}ll|llll|llll@{}}
\toprule
\multirow{2}{*}{\textbf{Model}}
& \multirow{2}{*}{\textbf{Effort}}
& \multicolumn{4}{c|}{\textit{URSA-expert-2026}}
& \multicolumn{4}{c}{\textit{URSA-drugs\&clinicals-2026}} \\
\cmidrule(lr){3-6}\cmidrule(lr){7-10}
& & \textbf{Solv-0} & \textbf{Solv-1} & \textbf{Solv-2} & \textbf{CC*}
& \textbf{Solv-0} & \textbf{Solv-1} & \textbf{Solv-2} & \textbf{CC*} \\
\midrule
\multicolumn{10}{c}{\textit{Proprietary Foundation Models}} \\
\midrule
Grok-4.1                & high   & 67 & 19 & 6 & 1.23 & 75 & 38 & 24 & 2.03 \\
Grok-4.3                & high   & 5 & 0 & 0 & 0.94 & 10 & 1 & 1 & 0.58 \\
Grok-4.5                & high   & 65 & 31 & 11 & 1.34 & 77 & 45 & 30 & 2.23 \\
Grok-4.6                & high   & 83 & 34 & 11 & 1.36 & 91 & 42 & 28 & 0.99 \\
Grok-4.7                & high   & 88 & 48 & 20 & 1.58 & 92 & 60 & 39 & 2.33 \\
Gemini 3.1 Pro          & high   & 54 & 29 & 12 & 1.45 & 70 & 47 & 38 & 2.42 \\
GPT 5.1                 & medium   & 16 & 1 & 1 & 0.34 & 11 & 1 & 0 & 0.38 \\
GPT 5.2                 & medium   & 12 & 1 & 0 & 0.67 & 34 & 10 & 9 & 1.37 \\
GPT 5.4                 & medium   & 20 & 1 & 0 & 0.67 & 29 & 9 & 5 & 1.43 \\
GPT 5.5                 & medium & 63 & 40 & 15 & 1.61 & 75 & 52 & 34 & 2.40 \\
GPT 5.6 Luna            & high   & 59 & 30 & 11 & 1.45 & 73 & 44 & 28 & 2.26 \\
GPT 5.6 Terra           & high   & 73 & 36 & 14 & 1.57 & 78 & 48 & 35 & 2.28 \\
GPT 5.6 Sol             & high   & 88 & 55 & \cellcolor{silver!40}28 & 1.74 & 87 & 64 & 40 & 2.40 \\
GPT 6 Luna              & high   & 86 & 44 & 20 & 1.64 & 83 & 56 & 36 & 2.29 \\
GPT 6 Sol               & high   & 94 & 62 & 25 & 1.83 & 95 & 62 & 39 & 2.32 \\
GPT 6 Astra             & medium & 86 & 63 & \cellcolor{bronze!40}27 & \underline{1.94} & 85 & 67 & 46 & 2.44 \\
GPT 6.1 Sol             & medium & 90 & \underline{70} & \cellcolor{gold!40}31 & 1.87 & 87 & 70 & 44 & 2.40 \\
Claude Sonnet 4.5       & high   & 35 & 3 & 1 & 0.97 & 59 & 13 & 7 & 1.49 \\
Claude Sonnet 4.6       & high   & 39 & 8 & 3 & 0.94 & 56 & 13 & 9 & 1.46 \\
Claude Opus 4.5         & high   & 45 & 6 & 3 & 1.02 & 72 & 21 & 18 & 1.72 \\
Claude Opus 4.6         & high   & 60 & 9 & 5 & 1.03 & 66 & 26 & 16 & 1.65 \\
Claude Opus 4.7         & high   & 51 & 15 & 8 & 1.23 & 60 & 30 & 23 & 1.99 \\
Claude Opus 4.8         & high   & 55 & 17 & 9 & 1.27 & 75 & 33 & 28 & 2.00 \\
Claude Opus 5           & high   & 52 & 38 & 17 & 1.66 & 70 & 53 & 39 & \underline{2.58} \\
\midrule
\multicolumn{10}{c}{\textit{Open-weight Foundation Models}} \\
\midrule
DeepSeek-V3.2           & high   & 11 & 1 & 0 & 0.84 & 15 & 2 & 1 & 1.08 \\
DeepSeek-V4.1 Flash     & high   & 69 & 29 & 14 & 1.35 & 78 & 47 & 32 & 2.23 \\
Qwen3.5                 & high   & 54 & 7 & 2 & 0.95 & 68 & 21 & 16 & 1.70 \\
Qwen3.8 Max             & xhigh  & 81 & 41 & 12 & 1.67 & 79 & 49 & 31 & 2.37 \\
Kimi K2.5               & high   & \underline{97} & 12 & 4 & 1.42 & \underline{98} & 16 & 10 & 1.79 \\
Kimi K3                 & max    & 55 & 18 & 8 & 1.21 & 76 & 35 & 29 & 2.00 \\
Kimi K3.1 Pro           & high   & 88 & 39 & 13 & 1.46 & 92 & 55 & 34 & 2.23 \\
GLM-5                   & high   & 66 & 5 & 2 & 1.00 & 71 & 13 & 9 & 1.46 \\
GLM-5.3                 & high   & 79 & 29 & 8 & 1.44 & 89 & 47 & 29 & 2.08 \\
\midrule
\multicolumn{10}{c}{\textit{Conventional Retrosynthesis Models}} \\
\midrule
Retro*                  & -- & 57 & 51 & 14 & 1.36 & 76 & 64 & 43 & 2.35 \\
AZF-MCTS                & -- & 65 & 57 & 24 & 1.58 & 81 & \underline{72} & \underline{51} & 2.26 \\
AZF-Retro*              & -- & 59 & 52 & 19 & 1.40 & 74 & 70 & \underline{51} & 2.31 \\
DMS-Wide                & -- & 85 & 18 & 12 & \textbf{2.02} & 87 & 53 & 41 & 2.55 \\
DMS-Flash               & -- & 72 & 21 & 14 & 1.89 & 78 & 52 & 42 & 2.53 \\
DMS-Explorer-XL         & -- & 40 & 15 & 10 & 1.34 & 54 & 40 & 34 & \textbf{3.17} \\
Retro*-0-LR             & -- & 53 & 37 & 17 & 1.36 & 58 & 38 & 25 & 1.85 \\
SynPlanner-MCTS         & -- & 14 & 10 & 6 & 1.84 & 38 & 37 & 28 & 2.43 \\
ASKCOS                  & -- & 23 & 18 & 12 & 1.39 & 60 & 50 & 43 & 2.39 \\
RetroChimera-MCTS       & -- & \textbf{99} & \textbf{71} & \cellcolor{silver!40}28 & 1.49 & \textbf{100} & \textbf{73} & \textbf{54} & 2.22 \\
DreamRetroer            & -- & 95 & 54 & 7 & 0.95 & 97 & 62 & 39 & 1.44 \\
SynLlama                & -- & 0 & 0 & 0 & 0.00 & 5 & 4 & 4 & 1.86 \\
TTLA                    & -- & 43 & 29 & 13 & 1.34 & 53 & 45 & 34 & 2.29 \\
\bottomrule
\end{tabular}

%% file: tables/appendix/u2p2.tex
\begin{tabular}{@{}l|llll|llll@{}}
\toprule
\multirow{2}{*}{\textbf{Model}}
 & \multicolumn{4}{c|}{\textit{URSA-expert-2026}}
 & \multicolumn{4}{c}{\textit{URSA-drugs\&clinicals-2026}} \\
\cmidrule(lr){2-5}\cmidrule(lr){6-9}
  & \textbf{Solv-0} & \textbf{Solv-1} & \textbf{Solv-2} & \textbf{CC*}
  & \textbf{Solv-0} & \textbf{Solv-1} & \textbf{Solv-2} & \textbf{CC*} \\
\midrule

\multicolumn{9}{c}{\textit{Proprietary Foundation Models}} \\
\midrule
Grok-4.1          & 67 & 29 & 10 & 1.54 & 75 & 43 & 31 & 2.76 \\
Grok-4.3          &  5 &  1 &  0 & 1.39 & 10 &  1 &  1 & 2.24 \\
Grok-4.5          & 65 & 43 & 25 & 1.81 & 77 & 61 & 46 & 3.20 \\
Grok-4.6          & 83 & 52 & 32 & 2.01 & 91 & 70 & 57 & 3.29 \\
Grok-4.7          & 88 & 71 & 46 & 2.12 & 92 & 80 & 65 & 3.28 \\
Gemini 3.1 Pro    & 54 & 37 & 24 & 1.84 & 70 & 56 & 47 & 3.44 \\
GPT 5.1           & 16 &  1 &  1 & 0.36 & 11 &  1 &  1 & 0.85 \\
GPT 5.2           & 12 &  2 &  0 & 0.90 & 34 & 13 &  8 & 2.05 \\
GPT 5.4           & 20 &  2 &  2 & 0.94 & 29 & 10 &  6 & 2.14 \\
GPT 5.5           & 63 & 46 & 29 & 2.01 & 75 & 59 & 47 & 3.19 \\
GPT 5.6 Sol       & 88 & 68 & 49 & 2.25 & 87 & 74 & 59 & 3.26 \\
GPT 5.6 Terra     & 73 & 50 & 28 & 1.92 & 78 & 53 & 47 & 2.97 \\
GPT 5.6 Luna      & 59 & 40 & 26 & 1.83 & 73 & 55 & 44 & 3.03 \\
GPT 6 Sol         & 94 & 82 & \cellcolor{gold!40}58 & 2.38 & 95 & 81 & 64 & 3.18 \\
GPT 6 Luna        & 86 & 55 & 38 & 2.16 & 83 & 72 & 54 & 3.11 \\
GPT 6 Astra       & 86 & 77 & 50 & 2.44 & 85 & 76 & 65 & 3.39 \\
GPT 6.1 Sol       & 90 & 82 & \cellcolor{silver!40}57 & 2.37 & 87 & 76 & 64 & 3.26 \\
Claude Sonnet 4.5 & 35 &  3 &  2 & 1.25 & 59 & 17 & 14 & 2.15 \\
Claude Sonnet 4.6 & 39 &  9 &  4 & 1.14 & 56 & 15 & 12 & 2.16 \\
Claude Opus 4.5   & 45 &  7 &  3 & 1.23 & 72 & 26 & 22 & 2.34 \\
Claude Opus 4.6   & 60 & 11 &  6 & 1.35 & 66 & 27 & 21 & 2.34 \\
Claude Opus 4.7   & 51 & 21 & 13 & 1.50 & 60 & 36 & 32 & 2.83 \\
Claude Opus 4.8   & 55 & 22 & 12 & 1.67 & 75 & 44 & 36 & 2.94 \\
Claude Opus 5     & 52 & 43 & 32 & 2.04 & 70 & 59 & 53 & 3.59 \\

\midrule
\multicolumn{9}{c}{\textit{Open-weight Foundation Models}} \\
\midrule
DeepSeek-V3.2       & 11 &  1 &  0 & 1.00 & 15 &  3 &  3 & 1.44 \\
DeepSeek-V4.1 Flash & 69 & 38 & 22 & 1.78 & 78 & 56 & 45 & 3.10 \\
Qwen3.5             & 54 &  9 &  4 & 1.20 & 68 & 24 & 19 & 2.25 \\
Qwen3.8 Max         & 81 & 55 & 31 & 2.07 & 79 & 60 & 47 & 3.30 \\
Kimi K2.5           & 97 & 23 & 10 & 1.90 & 98 & 28 & 20 & 2.52 \\
Kimi K3             & 55 & 31 & 16 & 1.46 & 76 & 44 & 39 & 2.87 \\
Kimi K3.1 Pro       & 88 & 61 & 33 & 1.93 & 92 & 74 & 58 & 3.15 \\
GLM-5               & 66 &  5 &  4 & 1.35 & 71 & 16 & 14 & 2.10 \\
GLM-5.3             & 79 & 53 & 27 & 1.91 & 89 & 62 & 47 & 2.92 \\

\midrule
\multicolumn{9}{c}{\textit{Conventional Retrosynthesis Models}} \\
\midrule
Retro*              & 57 & 53 & 21 & 1.59 & 76  & 67 & 51 & 3.04 \\
AZF-MCTS            & 65 & 59 & 39 & 1.88 & 81  & 73 & 60 & 2.82 \\
AZF-Retro*          & 59 & 55 & 34 & 1.69 & 74  & 72 & 60 & 2.92 \\
DMS-Wide            & 85 & 27 & 17 & 2.23 & 87  & 54 & 48 & 3.09 \\
DMS-Flash           & 72 & 22 & 16 & 2.12 & 78  & 53 & 47 & 3.04 \\
DMS-Explorer-XL     & 40 & 18 & 11 & 1.56 & 54  & 40 & 38 & 3.62 \\
Retro*-0-LR         & 53 & 42 & 27 & 1.64 & 58  & 43 & 31 & 2.77 \\
SynPlanner-MCTS     & 14 & 10 &  6 & 1.94 & 38  & 38 & 29 & 3.00 \\
ASKCOS              & 23 & 22 & 18 & 1.82 & 60  & 57 & 53 & 3.36 \\
RetroChimera-MCTS   & 99 & 83 & \cellcolor{bronze!40}51 & 1.87
                    & 100 & 83 & 72 & 3.52 \\
DreamRetroer        & 95 & 57 & 18 & 1.17 & 97  & 70 & 44 & 1.88 \\
SynLlama            &  0 &  0 &  0 & 0.00 &  5  &  5 &  4 & 2.29 \\
TTLA                & 43 & 31 & 21 & 1.63 & 53  & 45 & 42 & 3.05 \\
\bottomrule
\end{tabular}

%% file: tables/appendix/uspto-190.tex
\begin{tabular}{@{}l|llll|llll@{}}
\toprule
\multirow{2}{*}{\textbf{Model}}
 & \multicolumn{4}{c}{\textit{URSA-minor-1.3.0-U3}}
 & \multicolumn{4}{c}{\textit{URSA-major-1.3.0-U3P4}} \\
\cmidrule(lr){2-5}\cmidrule(lr){6-9}
  & \textbf{Solv-0} & \textbf{Solv-1} & \textbf{Solv-2} & \textbf{CC*}
  & \textbf{Solv-0} & \textbf{Solv-1} & \textbf{Solv-2} & \textbf{CC*} \\
\midrule

\multicolumn{9}{c}{\textit{Proprietary Foundation Models}} \\
\midrule
Grok 4.1                           & 50 & 14 & 6 & 1.79 & 50 & 20 & 9 & 2.04 \\
Grok 4.3                           & 7 & 0 & 0 & 0.07 & 7 & 1 & 0 & 1.88 \\
Gemini 3.1 Pro                     & 54 & 21 & 12 & 2.27 & 54 & 29 & 19 & 2.65 \\
GPT 5.2                            & 11 & 0 & 0 & 0.72 & 11 & 0 & 0 & 0.82 \\
GPT 5.4                            & 19 & 0 & 0 & 0.59 & 19 & 1 & 0 & 0.84 \\
GPT 5.5, medium                    & 72 & 20 & 6 & 1.97 & 72 & 26 & 12 & 2.24 \\
Claude Sonnet 4.5                  & 55 & 2 & 1 & 1.18 & 55 & 5 & 4 & 1.55 \\
Claude Sonnet 4.6                  & 48 & 3 & 2 & 1.14 & 48 & 5 & 3 & 1.45 \\
Claude Opus 4.5                    & 56 & 4 & 2 & 1.35 & 56 & 7 & 4 & 1.58 \\
Claude Opus 4.6                    & 61 & 5 & 3 & 1.33 & 61 & 11 & 6 & 1.68 \\
Claude Opus 4.7                    & 44 & 10 & 6 & 1.57 & 44 & 12 & 8 & 1.84 \\
Claude Opus 4.8                    & 58 & 10 & 5 & 1.61 & 58 & 15 & 7 & 2.02 \\

\midrule
\multicolumn{9}{c}{\textit{Conventional Retrosynthesis Models}} \\
\midrule
Retro*                             & 50 & 32 & 17 & 3.30 & 50 & 34 & 20 & 3.39 \\
DMS-Wide                           & 62 & 14 & 9 & 2.71 & 62 & 16 & 14 & 3.00 \\
DMS-Flash                          & 60 & 12 & 9 & 2.46 & 60 & 13 & 10 & 2.66 \\
DMS-Explorer-XL                    & 28 & 10 & 8 & 2.94 & 28 & 12 & 9 & 3.15 \\
Retro*-0-LR                        & 34 & 23 & 11 & 1.98 & 34 & 24 & 15 & 2.33 \\
ASKCOS                             & 15 & 11 & 6 & 2.54 & 15 & 14 & 10 & 2.83 \\
DreamRetroer                       & 91 & 33 & 11 & 1.87 & 91 & 36 & 15 & 2.02 \\
SynLlama                           & 1 & 1 & 0 & 0.00 & 1 & 1 & 0 & 0.00 \\
TTLA                               & 21 & 10 & 6 & 2.33 & 21 & 12 & 7 & 2.53 \\
SynPlanner-MCTS                    & 11 & 8 & 4 & 2.46 & 11 & 8 & 4 & 2.56 \\
AZF-MCTS                           & 34 & 25 & 11 & 2.79 & 34 & 26 & 18 & 3.02 \\
AZF-Retro*                     & 29 & 22 & 11 & 2.95 & 29 & 22 & 17 & 3.18 \\
RetroChimera-MCTS                  & 98 & 47 & 26 & 3.44 & 98 & 61 & 42 & 3.54 \\

\bottomrule
\end{tabular}

%% file: tables/appendix/providers.tex
\begin{tabular}{@{}llrrl@{}}
\toprule
\textbf{Model} & \textbf{Pricing source} & \multicolumn{2}{c}{\textbf{USD / 1M tokens}} & \textbf{Reference} \\
 & & \textbf{Input} & \textbf{Output} & \\
\midrule
\multicolumn{5}{c}{\textit{Proprietary Foundation Models}} \\
\midrule
Grok-4.7 & xAI & 2.00 & 6.00 & \href{https://docs.x.ai/developers/pricing}{xAI pricing} \\
Gemini 3.1 Pro & Google & 2.00 & 12.00 & \href{https://ai.google.dev/gemini-api/docs/pricing}{Gemini API pricing} \\
GPT 6 Astra & OpenAI & 10.00 & 50.00 & \href{https://developers.openai.com/api/docs/pricing}{OpenAI API pricing} \\
GPT 6.1 Sol & OpenAI & 2.00 & 10.00 & \href{https://developers.openai.com/api/docs/pricing}{OpenAI API pricing} \\
Claude Opus 5 & Anthropic & 5.00 & 25.00 & \href{https://platform.claude.com/docs/en/about-claude/pricing}{Claude API pricing} \\
\midrule
\multicolumn{5}{c}{\textit{Open-weight Foundation Models}} \\
\midrule
DeepSeek-V4.1 Flash & DeepSeek & 0.30 & 1.20 & \href{https://api-docs.deepseek.com/quick_start/pricing}{DeepSeek pricing} \\
Qwen3.8 Max & QwenCloud & 2.00 & 6.00 & \href{https://docs.qwencloud.com/developer-guides/getting-started/model-prices}{QwenCloud pricing} \\
Kimi K3.1 Pro & Moonshot AI & 2.00 & 10.00 & \href{https://platform.kimi.ai/docs/pricing/chat}{Kimi API pricing} \\
GLM-5.3 & Z.AI & 1.40 & 4.40 & \href{https://docs.z.ai/guides/overview/pricing}{Z.AI pricing} \\
\bottomrule
\end{tabular}

%% file: tables/appendix/costs.tex
\begin{tabular}{@{}l|rrr|rrr@{}}
\toprule
\multirow{2}{*}{\textbf{Model}}
 & \multicolumn{3}{c|}{\textit{URSA-expert-2026}}
 & \multicolumn{3}{c}{\textit{URSA-drugs\&clinicals-2026}} \\
\cmidrule(lr){2-4}\cmidrule(lr){5-7}
 & $N_{\mathrm{in}}$ & $N_{\mathrm{out}}$ & $C$
 & $N_{\mathrm{in}}$ & $N_{\mathrm{out}}$ & $C$ \\
\midrule
\multicolumn{7}{c}{\textit{Proprietary Foundation Models}} \\
\midrule
Grok-4.7        & 678 & 23877 & 144.62 & 678 & 23860 & 144.52 \\
GPT 6 Astra     & 469 & 3259  & 167.64 & 469 & 2836  & 146.49 \\
GPT 6.1 Sol     & 469 & 2667  & 27.60  & 474 & 2353  & 24.48 \\
Claude Opus 5   & 532 & 3011  & 77.93  & 532 & 2571  & 66.94 \\
\midrule
\multicolumn{7}{c}{\textit{Open-weight Foundation Models}} \\
\midrule
DeepSeek-V4.1 Flash & 468 & 29193 & 35.17  & 468 & 34793 & 41.89 \\
Qwen3.8 Max        & 523 & 31711 & 191.31 & 523 & 25513 & 154.12 \\
Kimi K3.1 Pro      & 524 & 6331  & 64.36  & 531 & 5795  & 59.01 \\
GLM-5.3            & 429 & 77050 & 339.62 & 426 & 75251 & 331.70 \\
\bottomrule
\end{tabular}

%% file: tables/appendix/costs_conv.tex
\begin{tabular}{@{}l|rrr|rrr@{}}
\toprule
\multirow{2}{*}{\textbf{Model}}
 & \multicolumn{3}{c|}{\textit{URSA-expert-2026}}
 & \multicolumn{3}{c}{\textit{URSA-drugs\&clinicals-2026}} \\
\cmidrule(lr){2-4}\cmidrule(lr){5-7}
 & $T$ & $H$ & $R$ & $T$ & $H$ & $R$ \\
\midrule
AZF-MCTS          & 18.1 & 0.1785 & 0.05 & 15.0 & 0.1785 & 0.04 \\
AZF-Retro*        & 66.6 & 0.1785 & 0.20 & 68.2 & 0.1785 & 0.20 \\
ASKCOS            & 51.7 & 0.7140 & 0.61 & 51.3 & 0.7140 & 0.61 \\
DMS-Explorer-XL   & 33.3 & 1.2900 & 0.72 & 36.4 & 1.2900 & 0.78 \\
DMS-Flash         & 152.2 & 1.2900 & 3.27 & 151.2 & 1.2900 & 3.25 \\
DMS-Wide          & 464.4 & 1.2900 & 9.98 & 467.1 & 1.2900 & 10.04 \\
DreamRetroer      & 4.6 & 0.1785 & 0.01 & 3.6 & 0.1785 & 0.01 \\
TTLA              & 4619.5 & 0.1785 & 13.74 & 4092.2 & 0.1785 & 12.17 \\
Retro*            & 16.4 & 0.1785 & 0.05 & 9.0 & 0.1785 & 0.03 \\
RetroChimera-MCTS & 28.05 & 3.0600 & 1.43 & 25.75 & 3.0600 & 1.31 \\
SynLlama          & 1.6 & 0.1785 & {<.01} & 2.7 & 0.1785 & 0.01 \\
Retro*-0-LR       & 12.1 & 0.1785 & 0.04 & 12.1 & 0.1785 & 0.04 \\
\bottomrule
\end{tabular}

%% file: text/appendix/da_dca.tex
\section{\texttt{URSA-drugs\&clinicals-2026} and \texttt{URSA-drugs\&clinicals\&agrochemicals-2026} Target Sets}
\label{sec:ursa_chemical_space}

\subsection{Datasets}

\texttt{URSA-drugs\&clinicals-2026} contains 100 approved drugs and clinical candidates selected as medicinally relevant retrosynthetic targets. \texttt{URSA-drugs\&clinicals\&agrochemicals-2026} extends the scope to 333 targets, comprising 300 drugs or clinical candidates and 33 agrochemicals. All structures were represented by canonical isomeric SMILES generated with RDKit, preserving explicitly specified stereochemistry.

For comparison with patent-derived chemistry, we used PaRoutes~2.0 \cite{paroutes_2022}. The \texttt{n1} and \texttt{n5} reference sets were analyzed separately. Each contains 10,000 unique target molecules. Only the published target SMILES were used; route intermediates, starting materials, stock molecules, and reaction records were excluded.

\subsection{BRICS fragment analysis}

BRICS fragments were generated using the default RDKit implementation,
\texttt{BRICS.BRICSDecompose(mol)}. BRICS attachment labels were retained when defining fragment-vocabulary identity. \texttt{URSA-drugs\&clinicals-2026} contained 320 unique labelled BRICS fragments, whereas \texttt{URSA-drugs\&clinicals\&agrochemicals-2026} contained 869.

To control for dataset size, we drew 10,000 independent molecule-level samples from each PaRoutes set, using sample sizes of $N=100$ and $N=333$. Sampling was performed without replacement using a fixed random seed. The observed URSA vocabulary size was compared with the resulting PaRoutes reference distribution.

\begin{table}[H]
\centering
\setlength{\tabcolsep}{5pt}
\begin{tabular}{@{}lrrr@{}}
\toprule
PaRoutes set & URSA fragments & PaRoutes mean (95\% interval) & Ratio \\
\midrule
\multicolumn{4}{@{}l}{\texttt{URSA-drugs\&clinicals-2026} ($N=100$)} \\
\midrule
n1 & 320 & 248.0 (231--265) & 1.290 \\
n5 & 320 & 259.4 (243--277) & 1.233 \\
\midrule
\addlinespace[3pt]
\multicolumn{4}{@{}l}{\texttt{URSA-drugs\&clinicals\&agrochemicals-2026} ($N=333$)} \\
\midrule
n1 & 869 & 637.6 (612--664) & 1.363 \\
n5 & 869 & 661.3 (634--688) & 1.314 \\
\bottomrule
\end{tabular}
\caption{BRICS vocabulary richness relative to size-matched PaRoutes target samples. Intervals span the 2.5th to 97.5th percentiles across 10,000 resamples.}
\label{tab:brics_richness}
\end{table}

\texttt{URSA-drugs\&clinicals-2026} therefore contained 23--29\% more BRICS fragments than an equally sized PaRoutes sample, whereas \texttt{URSA-drugs\&clinicals\&agrochemicals-2026} contained 31--36\% more. When \texttt{URSA-drugs\&clinicals\&agrochemicals-2026} was downsampled to 100 targets, its mean vocabulary was 328.2 fragments (95\% interval: 311--345), comparable to the 320 fragments observed for \texttt{URSA-drugs\&clinicals-2026}. Thus, the larger total vocabulary of \texttt{URSA-drugs\&clinicals\&agrochemicals-2026} is primarily attributable to its increased number of targets and improved coverage of the fragment-space long tail, rather than a substantially greater fragment diversity per target.

\subsection{Internal structural diversity}

Pairwise molecular similarity was evaluated using a maximum-common-substructure (MCS) Tanimoto coefficient,

\begin{equation}
T_{\mathrm{MCS}}(A,B) =
\frac{N_{\mathrm{MCS}}}
{N_A + N_B - N_{\mathrm{MCS}}},
\end{equation}

where $N_A$ and $N_B$ are the heavy-atom counts of molecules $A$ and $B$, and $N_{\mathrm{MCS}}$ is the number of heavy atoms in their MCS. Atoms were matched by element and bonds by order; ring atoms were constrained to match ring atoms and only complete rings were accepted.

Among the 4,950 molecular pairs in \texttt{URSA-drugs\&clinicals-2026}, one pair (0.020\%) had $T_{\mathrm{MCS}}\geq0.4$ and also exceeded 0.5. Among the 55,278 pairs in \texttt{URSA-drugs\&clinicals\&agrochemicals-2026}, three pairs (0.005\%) reached $T_{\mathrm{MCS}}\geq0.4$, and none reached 0.5. These results indicate low structural redundancy within both target sets.

\subsection{Coverage of PaRoutes fragment space}

For structural fragment comparisons, BRICS dummy atoms were replaced by hydrogen before MCS calculation. For each PaRoutes fragment $p$, coverage at threshold $\tau$ was defined as

\begin{equation}
C_{\tau}(p) =
\mathbb{I}
\left[
\max_{u \in V_{\mathrm{URSA}}}
T_{\mathrm{MCS}}(p,u) \geq \tau
\right],
\end{equation}

where $V_{\mathrm{URSA}}$ is the corresponding URSA fragment vocabulary. We report both unweighted vocabulary coverage and coverage weighted by the number of PaRoutes target molecules containing each fragment.

\begin{table}[H]
\centering
\begin{tabular}{@{}lcc@{}}
\toprule
PaRoutes set & $T_{\mathrm{MCS}}\geq 0.4$ & $T_{\mathrm{MCS}}\geq 0.6$ \\
\midrule
\multicolumn{3}{@{}l}{\texttt{URSA-drugs\&clinicals-2026} ($N=100$)} \\
\midrule
n1 & 94.18\% (98.08\%) & 77.65\% (93.99\%) \\
n5 & 94.39\% (98.22\%) & 79.61\% (94.65\%) \\
\addlinespace[3pt]
\midrule
\multicolumn{3}{@{}l}{\texttt{URSA-drugs\&clinicals\&agrochemicals-2026} ($N=333$)} \\
\midrule
n1 & 96.97\% (99.24\%) & 84.91\% (96.35\%) \\
n5 & 97.21\% (99.38\%) & 86.01\% (96.87\%) \\
\bottomrule
\end{tabular}
\caption{Coverage of PaRoutes target-fragment space by the URSA datasets. Target-frequency-weighted coverage is shown in parentheses.}
\label{tab:paroutes_coverage}
\end{table}

Both URSA datasets therefore cover most of the BRICS-derived synthon space represented by PaRoutes targets. The addition of drugs, clinical candidates, and agrochemicals in \texttt{URSA-drugs\&clinicals\&agrochemicals-2026} improves unweighted coverage by approximately 2.8 percentage points at $T_{\mathrm{MCS}}\geq0.4$ and by 6.4--7.3 percentage points at the stricter threshold of 0.6. The high frequency-weighted coverage indicates that nearly all commonly occurring PaRoutes fragments are represented by structurally similar fragments in URSA.

These results support broad coverage of patent-derived structural and synthon space. They do not, by themselves, establish coverage of reaction classes, bond-forming transformations, synthetic feasibility, or route quality.

\subsection{\texttt{URSA-drugs\&clinicals\&agrochemicals-2026} Benchmarking Results}
\label{app:dca_results}

We additionally evaluated selected LLMs and conventional CASP tools on
\texttt{URSA-drugs\&clinicals\&agrochemicals-2026}, an extended benchmark
containing 333 target molecules. The results are reported in
\autoref{tab:dca_benchmark}.

\begin{table}[h]
\centering
\begin{tabular}{@{}lcccc@{}}
\toprule
\textbf{Model} & \textbf{Solv-0} & \textbf{Solv-1}
               & \textbf{Solv-2} & \textbf{CC*} \\
\midrule
\multicolumn{5}{c}{\textit{Proprietary Foundation Models}} \\
\midrule
Gemini 3.1 Pro      & 64 & 39 & 30 & 2.18 \\
GPT 5.5             & 73 & 42 & 26 & 2.17 \\
Claude Opus 4.7     & 53 & 21 & 15 & 1.66 \\
\midrule
\multicolumn{5}{c}{\textit{Conventional Retrosynthesis Models}} \\
\midrule
RetroChimera-MCTS   & 98 & 67 & 45 & 1.99 \\
AZF-MCTS            & 68 & 60 & 44 & 2.21 \\
AZF-Retro*          & 65 & 57 & 44 & 2.34 \\
Retro*-0-LR         & 69 & 37 & 25 & 1.87 \\
Retro*              & 74 & 47 & 28 & 2.08 \\
\bottomrule
\end{tabular}
\caption{Solvability benchmark results on the 333-target
\texttt{URSA-drugs\&clinicals\&agrochemicals-2026} set. Solv-0,
Solv-1, and Solv-2 are percentages of target molecules. \textbf{CC*}
is the average ChemCensor score per step along the best route,
averaged across targets (see \autoref{eq:chemcensor}).}
\label{tab:dca_benchmark}
\end{table}

%% file: text/appendix/MSTS_templates.tex
\clearpage
\section{Templates Used for LLMs Benchmarking}
\label{app:msrs_templates}

To formulate retrosynthesis tasks in natural language, we use a set of templates. We wrap each query into 10 different templates, forming the final data representation. These templates in Jinja2 format are listed below. We used the SMILES molecular format by default.

\begin{windowbox}[width=\linewidth]{Templates for the multistep retrosynthesis task}
\footnotesize
\begin{fieldbox}{Template 1}
Please propose a plausible synthetic route for the target molecule \textless\textcolor{blue}{\{\{input\_format\}\}}\textgreater\textcolor{blue}{\{\{query\}\}}\textless/\textcolor{blue}{\{\{input\_format\}\}}\textgreater. The route must converge to commercially available starting materials (purchasable building blocks).
\end{fieldbox}

\begin{fieldbox}{Template 2}
Please propose a stepwise retrosynthetic plan for the target molecule \textless\textcolor{blue}{\{\{input\_format\}\}}\textgreater\textcolor{blue}{\{\{query\}\}}\textless/\textcolor{blue}{\{\{input\_format\}\}}\textgreater{} in which the final precursors are limited to commercially supplied reagents and building blocks.
\end{fieldbox}

\begin{fieldbox}{Template 3}
Please propose a convergent retrosynthetic pathway for the target molecule \textless\textcolor{blue}{\{\{input\_format\}\}}\textgreater\textcolor{blue}{\{\{query\}\}}\textless/\textcolor{blue}{\{\{input\_format\}\}}\textgreater. It is mandatory that the route ends on commercially available starting materials, not bespoke intermediates.
\end{fieldbox}

\begin{fieldbox}{Template 4}
Design the entire synthetic route to reach the target molecule \textless\textcolor{blue}{\{\{input\_format\}\}}\textgreater\textcolor{blue}{\{\{query\}\}}\textless/\textcolor{blue}{\{\{input\_format\}\}}\textgreater{} from catalog chemicals. Explicitly ensure the retrosynthesis bottoms out in commercially accessible building blocks.
\end{fieldbox}

\begin{fieldbox}{Template 5}
For the target molecule \textless\textcolor{blue}{\{\{input\_format\}\}}\textgreater\textcolor{blue}{\{\{query\}\}}\textless/\textcolor{blue}{\{\{input\_format\}\}}\textgreater, outline a retrosynthetic analysis and propose a synthetic route. The route must terminate in reagents and building blocks that are commercially available from standard suppliers.
\end{fieldbox}

\begin{fieldbox}{Template 6}
Please propose a sequence of retrosynthetic disconnections that constitutes a valid synthetic route for the target molecule \textless\textcolor{blue}{\{\{input\_format\}\}}\textgreater\textcolor{blue}{\{\{query\}\}}\textless/\textcolor{blue}{\{\{input\_format\}\}}\textgreater. The route must lead to commercially available starting materials.
\end{fieldbox}

\begin{fieldbox}{Template 7}
Please perform retrosynthetic analysis and propose a synthetic route for the molecule \textless\textcolor{blue}{\{\{input\_format\}\}}\textgreater\textcolor{blue}{\{\{query\}\}}\textless/\textcolor{blue}{\{\{input\_format\}\}}\textgreater. The route is acceptable only if every terminal precursor can be sourced as a commercially available building block.
\end{fieldbox}

\begin{fieldbox}{Template 8}
Please propose a practical multi-step retrosynthetic plan for the target molecule \textless\textcolor{blue}{\{\{input\_format\}\}}\textgreater\textcolor{blue}{\{\{query\}\}}\textless/\textcolor{blue}{\{\{input\_format\}\}}\textgreater, in which all reactions ultimately rest on commercially catalogued building blocks as the inputs.
\end{fieldbox}

\begin{fieldbox}{Template 9}
Please propose the retrosynthetic pathway from the target molecule \textless\textcolor{blue}{\{\{input\_format\}\}}\textgreater\textcolor{blue}{\{\{query\}\}}\textless/\textcolor{blue}{\{\{input\_format\}\}}\textgreater{} down to commercially available building blocks.
\end{fieldbox}

\begin{fieldbox}{Template 10}
Please propose a retrosynthetic scheme for the target molecule \textless\textcolor{blue}{\{\{input\_format\}\}}\textgreater\textcolor{blue}{\{\{query\}\}}\textless/\textcolor{blue}{\{\{input\_format\}\}}\textgreater. The route must lead to commercially purchasable reagents and building blocks.
\end{fieldbox}

\end{windowbox}

%% file: text/appendix/uspto-190-ctitics.tex
\clearpage
\section{The structural diversity and relevance analysis of USPTO-190 set} 
\label{app:uspto_190_critics}

\begin{figure*}[th]
  \centering
  \includegraphics[
    %page=1,
    width=\linewidth,
    trim=0 0 0 0,
    clip
  ]{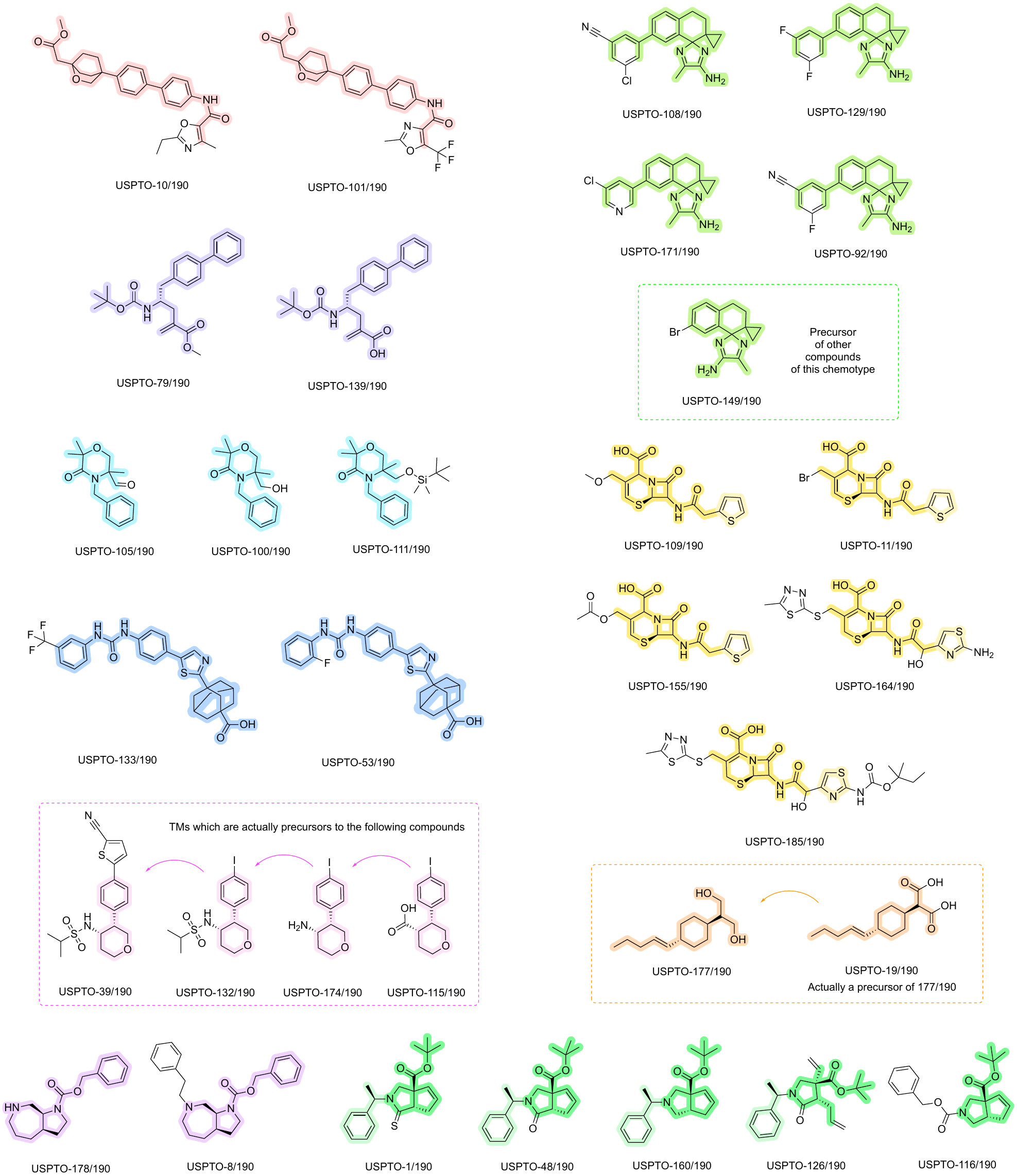}
  \caption{
Clusters of similar molecules from \texttt{USPTO-190}. Maximum common substructure (MCS) is highlighted independently for each cluster. 
}
  \label{fig:uspto_1}
\end{figure*}

\begin{figure*}[th]
  \centering
  \includegraphics[
    %page=1,
    width=\linewidth,
    trim=0 0 0 0,
    clip
  ]{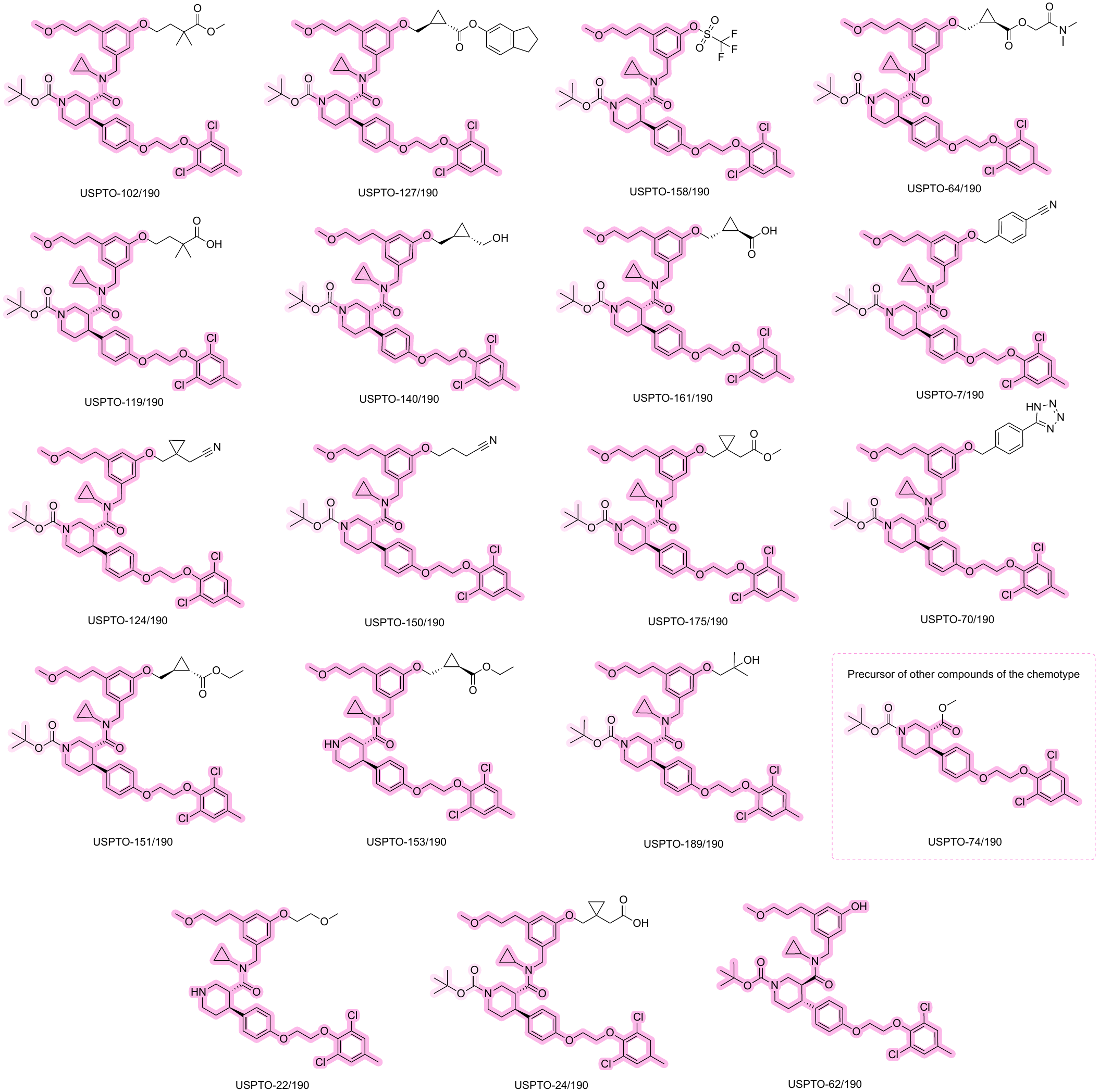}
  \caption{
The biggest cluster of highly similar structures from \texttt{USPTO-190}. Maximum common substructure (MCS) is highlighted for the cluster
}
  \label{fig:uspto_2}
\end{figure*}

The close structural analysis of \texttt{USPTO-190} set reveals its substantial diversity flaws. Many structures of \texttt{USPTO-190} can be easily clustered in pairs of strikingly similar entries \autoref{fig:uspto_1}, while the most prominent cluster consists of 19 structures \autoref{fig:uspto_2}, which are different only from the perspective of small peripheral substructures. The smaller clusters consist of 3-5 structures. 

Another finding about structural clusters in \texttt{USPTO-190} reflects another aspect of the unsystematic approach that was adopted during the nomination of compounds to this benchmarking set. Many structures are actually the precursors of other compounds of this set or even represent series of precursors (like USPTO-132/190, USPTO-174/190 and USPTO-115/190), which is unreasonable, since precursors should be present in the synthetic routes of true target molecules, making the inclusion of precursors partially redundant.

Some TMs from \texttt{USPTO-190} are not relevant to the current trends of medicinal chemistry design like the highlighted cephalosporins (yellow, \autoref{fig:uspto_1}), nucleoside analogs (USPTO-118/190, USPTO-144/190) and steroid (USPTO-75/190) that can be found in the original set. 

These highlighted flaws of the \texttt{USPTO-190} set are aimed to stress the attention of the community to the quality of benchmarking sets. Prior to adoption, the benchmarking sets should be rigorously analyzed to avoid potential biases in the evaluation procedures due to structural biases caused by limited diversity of the dataset.

%% file: text/appendix/subtree_gen_example.tex
\section{Route-Level Solv-2 Evaluation and Subtree Generation}
\label{app:subtree_gen_example}

\subsection{Conjunctive route validity}
\label{app:route_level_solv2}

Let $\mathcal{R}_t$ be the candidate routes for target $t$ after
standardization and subtree generation. For a candidate route $r$,
let $I(r)$ indicate that its molecular graphs and directed-acyclic
connectivity pass the route-integrity checks, $B(r)$ indicate that
all terminal molecules belong to the specified starting-material
stock, and $P_2(e)$ indicate that reaction step $e$ passes the
ChemCensor-based plausibility criterion. Route-level Solv-2 is

\begin{equation}
\mathrm{Solv}_2(r)
=
I(r)\,B(r)\prod_{e\in E(r)}P_2(e),
\end{equation}

where $E(r)$ is the set of reaction steps in $r$. Thus, one failed
required step makes the entire candidate route Solv-2 invalid.
A target passes Solv-2 if at least one of its candidate routes passes:
$\mathrm{Solv}_2(t)=\max_{r\in\mathcal{R}_t}\mathrm{Solv}_2(r)$.

The mean ChemCensor score of a route,
$\overline{\mathrm{CC}}(r)=|E(r)|^{-1}
\sum_{e\in E(r)}\mathrm{CC}_e$, serves a different purpose.
For example, assuming both routes pass the integrity and stock checks,
a route with step scores $(5,5,0)$ has
$\overline{\mathrm{CC}}(r)=3.33$ but fails Solv-2.
A route with scores $(2,2,2)$ has a lower mean score of $2.00$
but passes Solv-2. URSA therefore prefers the latter route.
The reported CC* averages selected route scores across targets
(see \autoref{eq:chemcensor}); it is not the route-level
Solv-2 decision rule.

\subsection{Subtree generation}

\begin{figure*}[th]
  \centering
  \includegraphics[
    %page=1,
    width=\linewidth,
    trim=30 70 25 70,
    clip
  ]{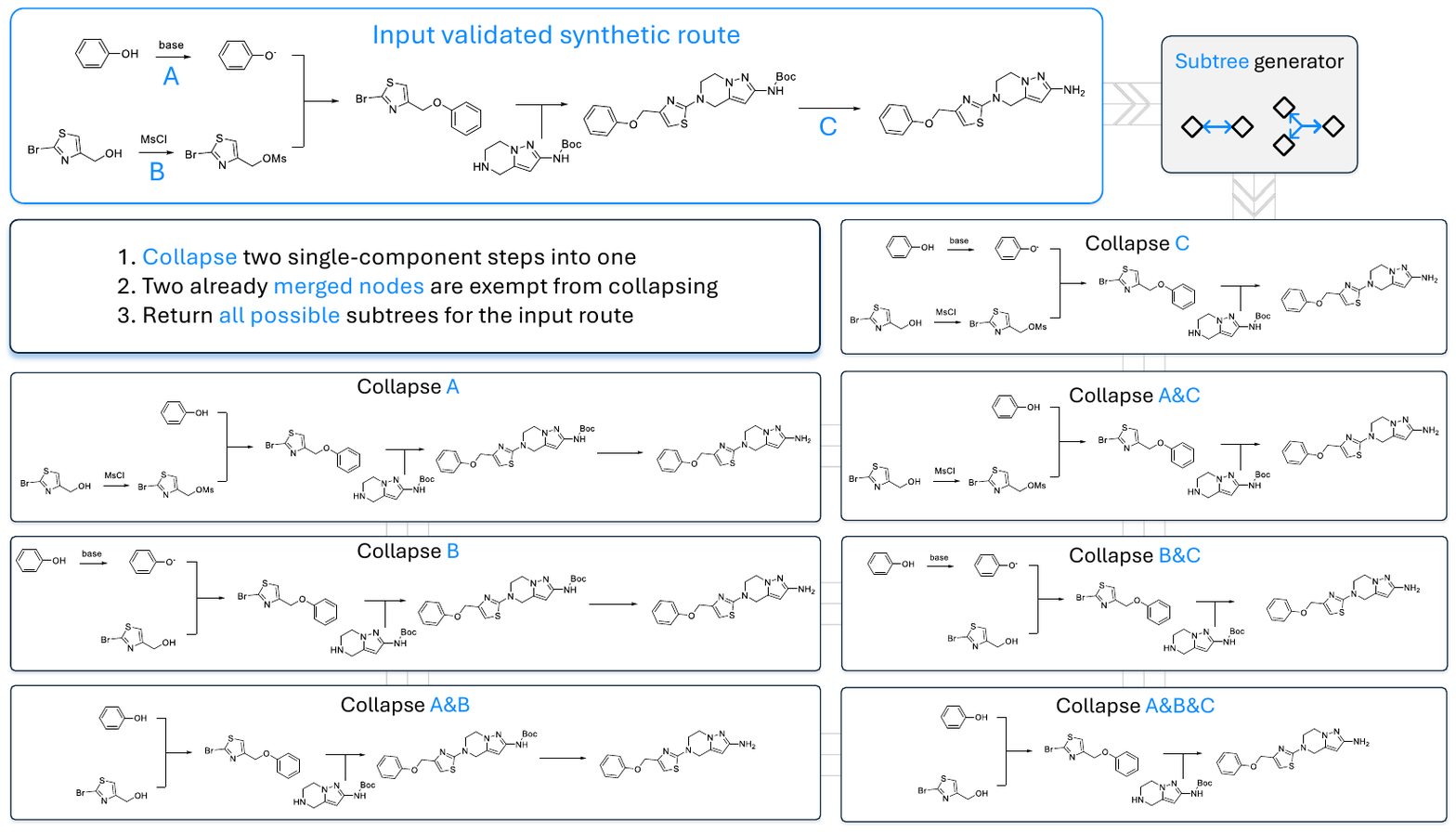}
  \caption{
Example of the subtree generation process.
}
  \label{fig:subtrees}
\end{figure*}

URSA evaluates each generated subroute as a complete candidate route
under the same conjunctive criterion. Collapsing eligible
one-component steps and merging adjacent reaction nodes provide
alternative representations of transformations that may be encoded
at different levels of granularity in a planner output and a
precedent database. ChemCensor then evaluates each reaction in the
candidate subroute using its complete reactant and product structures,
including the functional-group context present at that step.
The example route schema and its generated subroutes are shown in
\autoref{fig:subtrees}.

%% file: text/appendix/ChemCensor.tex
\newpage
\section{ChemCensor Scoring}
\label{app:chemcensor_rc}

\begin{figure*}[ht]
  \centering
  \includegraphics[
    width=\linewidth,
    clip,
    trim=20 160 300 70,
  ]{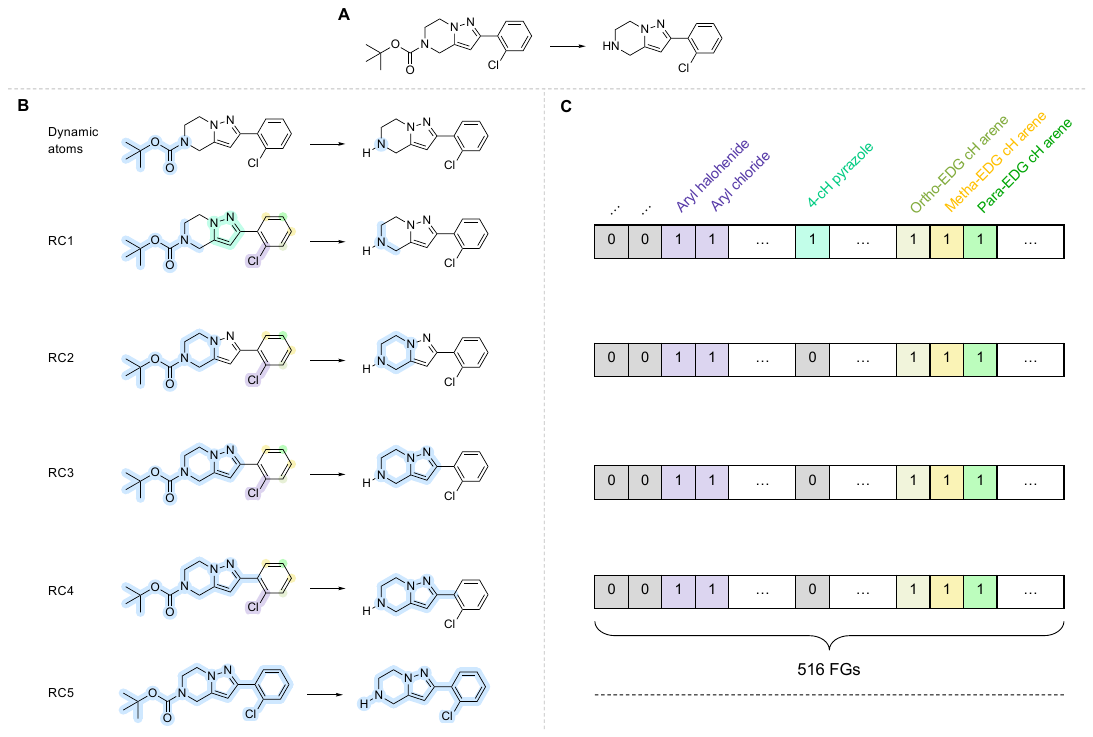}
  \caption{ChemCensor functionalities. \textbf{A}. The input reaction. \textbf{B}. Process of reaction center extraction with different RC\textbf{N} highlighted in blue. \textbf{C}. The respective FG signatures annotated for the extracted centers.
}
  \label{fig:RCFG}
\end{figure*}

ChemCensor estimates how closely a single reaction resembles a reference corpus of synthetic precedents and uses that resemblance as a proxy for chemical plausibility. Besides precedent estimation, it also performs basic checks like correct reaction syntax and SMILES validity. The pipeline standardizes the input, validates SMILES, and maps atoms between reactants and products (RxnMapper \cite{Schwaller2021rxnmapper} was used). Reactive atoms (atoms related to a dynamic part) are inferred from the mapped transform. Next, the processed reaction undergoes rigorous stereo validation: if there are any allowed stereo inversions (e.g., SN\textsubscript{2} reactions), legit stereo-center emergence (a procedure of the resolution of stereoisomers), or stereo errors (inversions in the static part of the reaction). Also, to properly render regioselectivity of electrophilic aromatic substitutions, the separate module, SeAr annotator, determines if the reaction belongs to S\textsubscript{e}Ar class and annotates the functional group context surrounding the C-H center. Once processing is done, the reaction center extractor yields a nested family of centers (RC1–RC4) with increasing peripheral context. Each center is additionally annotated with non-participating functional groups which are found outside the reactive part of the transformation. Finally, the scorer compares the extracted centers with centers prepared in the same manner for the reference database of synthetic precedents. It accepts a center if its signature is contained in the reference (sub-signature match), with higher levels yielding higher scores which implies the higher chemical plausibility. An exact canonical reaction hit can yield the top score. Additionally, S\textsubscript{e}Ar scoring includes checking the S\textsubscript{e}Ar context stored with each precedent reaction. The details on center extraction and FG annotation are provided below.

In the current version, there are five levels of reaction center annotation, denoted RC1, RC2, RC3, RC4 and RC5 corresponding to confidence levels and ChemCensor Scores 1, 2, 3, 4 and 5, respectively (see \autoref{fig:RCFG} \textbf{A)}. The RC\textbf{N} is the representation of an RC, where \textbf{N} shows the portion of the structural context of an RC; the higher the \textbf{N}, the higher the context size (see \autoref{fig:RCFG} \textbf{B}):

\begin{itemize}
    \item RC1 captures the minimal local environment of RC with the inclusion of dynamic atoms and atoms at topological distance of 1 from them. Also, atoms from predefined FGs (e.g., carbonyl group, nitrile, etc.) are included when at least one atom of a FG is dynamic.
    \item  RC2 includes atoms from RC1 with additional incorporation of the primary ring context (if present), stereochemical surroundings, and/or the atoms at topological distance of 2 around dynamic atoms.
    \item The next level, RC3, extends the atoms of RC2 by the addition of a fused-ring context around (if present) or the atoms at topological distance of 3 around dynamic atoms.
    \item Finally, RC4 is further extended by adding substituent context to aromatic systems and completing the predefined FGs, e.g., carbonyl group, nitrile, etc., or the atoms at topological distance of 4 from reactive atoms.
\end{itemize}

Scoring for S\textsubscript{e}Ar reactions also implies comparison of S\textsubscript{e}Ar functional group context surrounded by a C-H center with the context of the same center compiled for reported S\textsubscript{e}Ar reaction examples.

After extraction, each reaction center is annotated with functional groups outside the reaction center part. ChemCensor uses a curated database of 516 functional group patterns to cover basic chemical entities relevant to organic chemistry. Once matched, a non-reacting FG is considered tolerable during transformation if all the atoms of a FG are not overlapped with RC atoms. Besides that, FG signatures of centers derived from S\textsubscript{e}Ar reactions are additionally annotated with patterns describing different C-H patterns (e.g., presence of ortho-/para-EWG or EDG groups) which are normally not checked for centers from non-S\textsubscript{e}Ar transformations. This helps better evaluate regioselectivity of S\textsubscript{e}Ar reactions without affecting other reaction classes by poorly reactive C-H patterns.

During the processing of the corpus of reference reaction data, an FG signature is prepared to store the cumulative ensemble of non-participating groups found for all reaction examples associated with the given RC. To expedite the handling and searching, each reaction center is treated as a canonicalized SMARTS string describing reacting patterns for both left and right sides of the reaction. The annotated FG signatures are encoded as numpy binary arrays in the form of 516-bit signatures (\autoref{fig:RCFG} \textbf{C}). Reaction centers and FG signatures prepared for the dataset of synthetic precedents are stored in a SQLite database together with the reaction examples from which they originate.

%% file: text/appendix/Synthegy.tex
\clearpage
\section{Selected flaws of the Synthegy approach} 
\label{app:synthefy_critics}
While suggesting an interesting Synthegy framework to use LLMs as evaluators of retrosynthetic models' outcomes, the respective study \cite{Schwaller2026-synthegy} contains striking flaws of LLMs failing to judge correctly on chemical synthesis matters. 

\begin{figure*}[th]
  \centering
  \includegraphics[
    %page=1,
    width=\linewidth,
    trim=10 1 20 5,
    clip
  ]{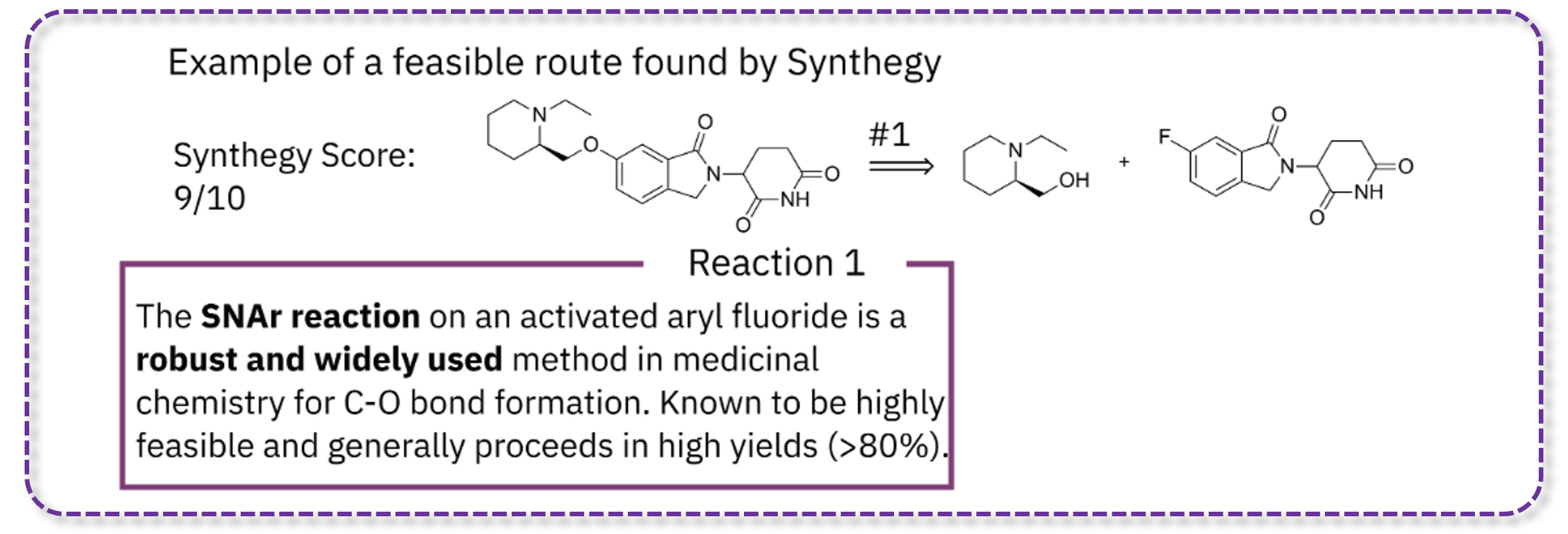}
  \caption{
Reaction erroneously understood as an S\textsubscript{n}Ar reaction by the Synthegy framework.
}
  \label{fig:synthegy_1}
\end{figure*}

\textbf{Example 1}. \autoref{fig:synthegy_1} presents such a case (taken from Fig. 3B of the original paper). As we can see Synthegy "thinks" that the provided reaction No.1 is a good and plausible one, since it represents the well-described S\textsubscript{n}Ar reaction. 

\begin{figure*}[th]
  \centering
  \includegraphics[
    %page=1,
    width=\linewidth,
    trim=10 1 20 5,
    clip
  ]{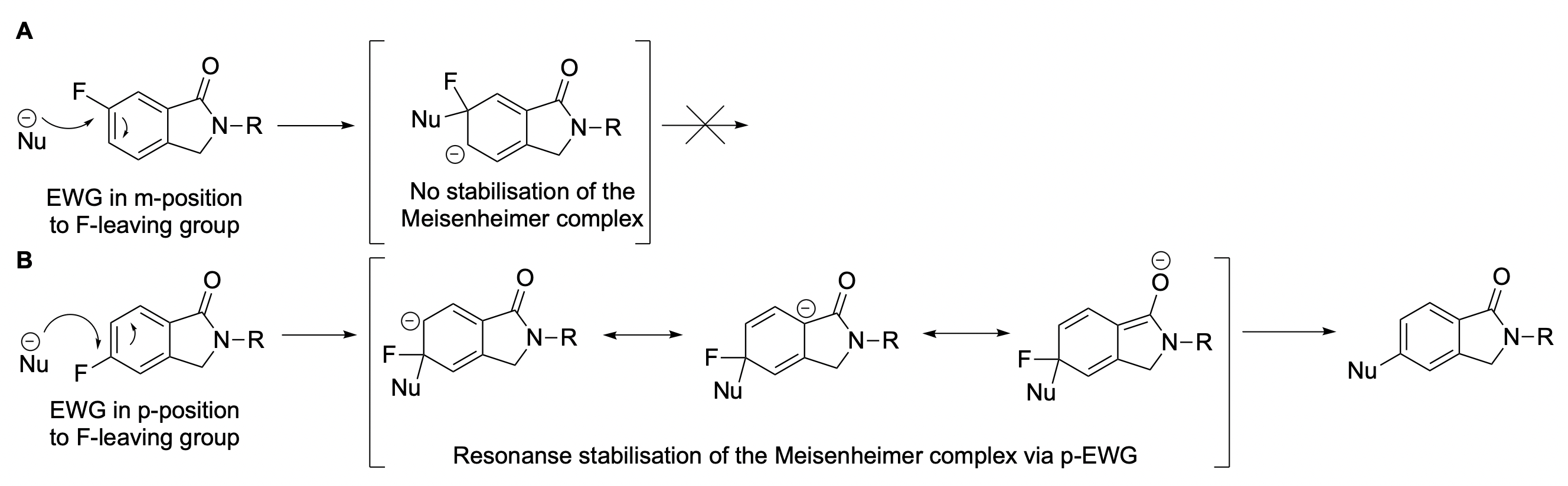}
  \caption{
Mechanism supporting the correct interpretation of the highlighted reaction.
}
  \label{fig:mechanism}
\end{figure*}

However, this is a classic example of LLMs misinterpreting structural information. The given aryl fluoride has the EWG (amide) in m-position, while to be readily substituted the EWG should be placed either in p- or in o-position to the leaving fluoride group (see \autoref{fig:mechanism}). This can be easily justified via the mechanism of S\textsubscript{n}Ar reaction that requires stabilization of the transition mode (Meisenheimer complex) that can be supported by o- or p- placed EWG. 

\begin{figure*}[th]
  \centering
  \includegraphics[
    %page=1,
    width=\linewidth,
    trim=10 1 20 5,
    clip
  ]{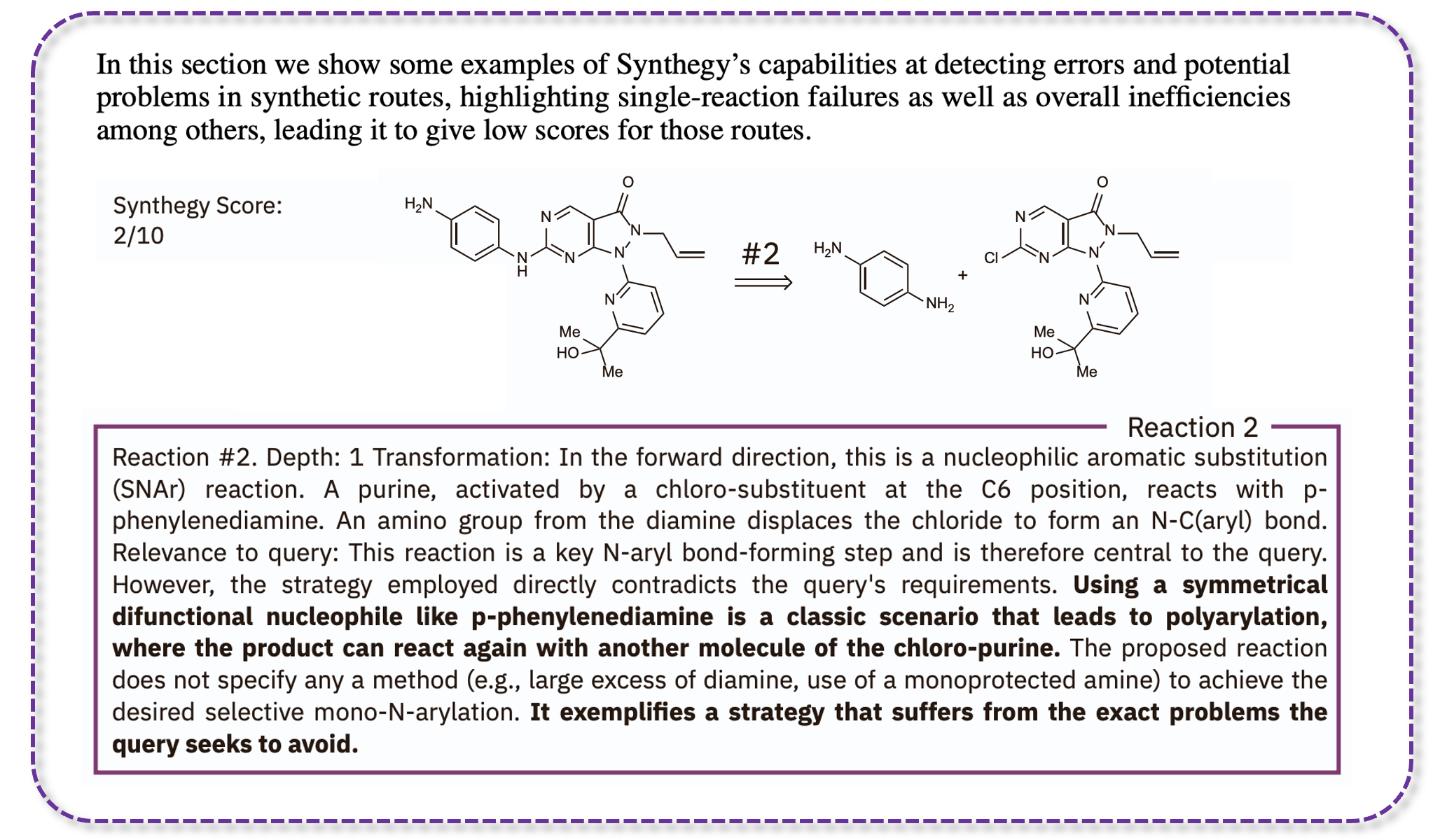}
  \caption{
Reaction unfairly penalized by Synthegy.
}
  \label{fig:synthegy_2}
\end{figure*}

\textbf{Example 2}. In contrast to \textbf{Example 1}, this (see \autoref{fig:synthegy_2}) case presents an erroneous penalization of feasible reaction by the Synthegy workflow. This case is taken from \textbf{SI H} Supplemental Note 8 Section of the original paper. Synthegy 'thinks' that this reaction can lead to polyarylation and does not suggest it for usage. 

\begin{figure*}[th]
  \centering
  \includegraphics[
    %page=1,
    width=\linewidth,
    trim=10 1 20 5,
    clip
  ]{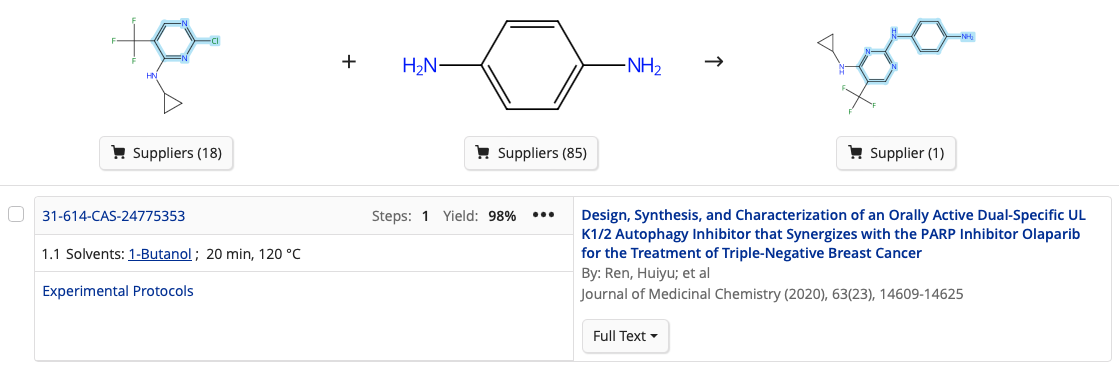}
  \caption{
An example from literature supporting the reaction penalized by Synthegy.
}
  \label{fig:scifinder_1}
\end{figure*}

However, a quick check in SciFinder \cite{scifinder} for the same reaction center and 2,4-diaminobenzene gives us a perfectly matching example (see \autoref{fig:scifinder_1}) from the Journal of Medicinal Chemistry \cite{Ren2020-el} of reaction with a high yield. While the 2x excess (not a large one) of cheap 2,4-diaminobenzene is involved, the practical utility of the reaction could not be diminished.

%% file: text/appendix/ranking_sensetivity.tex
\clearpage